%% file: main.tex
\documentclass[10pt]{article}

\usepackage{lmodern}
\usepackage[T1]{fontenc}
\usepackage[utf8]{inputenc}
\input{glyphtounicode}
\usepackage{microtype}
\microtypesetup{expansion=false}
\usepackage{placeins}
\usepackage{hyphenat}
\usepackage{setspace}
\usepackage{parskip}
\usepackage[english]{babel}
\usepackage{etoolbox}
\DisableLigatures[f]{family=sf*}

\usepackage[
  top=2.25cm, bottom=2.5cm,
  left=2.5cm, right=2.5cm,
  columnsep=0.65cm
]{geometry}

\usepackage{amsmath,amssymb,amsfonts,amsthm}
\usepackage{bm}
\usepackage{mathtools}

\usepackage{graphicx}
\usepackage{float}
\usepackage[dvipsnames,svgnames,x11names]{xcolor}
\definecolor{metablue}{HTML}{0064E0}
\usepackage{tikz}
\usepackage{pgfplots}
\pgfplotsset{compat=1.18}
\usetikzlibrary{
  arrows.meta, positioning, shapes.geometric,
  fit, backgrounds, patterns, calc,
  decorations.pathreplacing, decorations.markings
}
\usepgfplotslibrary{fillbetween}

\usepackage{booktabs}
\usepackage{multirow}
\usepackage{makecell}
\usepackage{tabularx}
\usepackage{array}
\usepackage{colortbl}

\usepackage[ruled,vlined,linesnumbered]{algorithm2e}

\usepackage[hang,flushmargin]{footmisc}

\usepackage{hyperref}
\usepackage{cleveref}
\hypersetup{
  colorlinks=true,
  linkcolor=metablue,
  citecolor=metablue,
  urlcolor=metablue,
  pdftitle={PRIME W0: Physical Priors from World Models for Embodied Manipulation},
  pdfauthor={Huabin Peng et al.}
}

\usepackage{caption}
\usepackage{subcaption}
\usepackage{enumitem}

\usepackage{titlesec}
\titleformat*{\section}{\Large\sffamily\bfseries}
\titleformat*{\subsection}{\large\sffamily\bfseries}
\titleformat*{\subsubsection}{\normalsize\sffamily\bfseries}
\DeclareTextFontCommand{\textbf}{\bfseries\sffamily}

\usepackage{fancyhdr}
\fancypagestyle{firststyle}{
  \fancyhf{}
  \setlength{\headwidth}{\textwidth}

  \fancyhead[R]{\includegraphics[height=0.5cm]{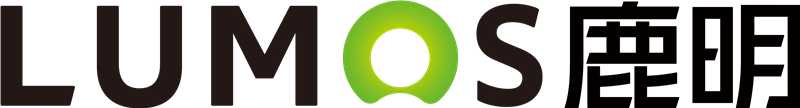}}
  \fancyfoot[C]{\small\thepage}
  \fancyfoot[L]{\fontsize{8}{9}\selectfont
    $^{*}$Correspondence: \texttt{chenliqing@lumosbot.tech}}
}
\fancypagestyle{plain}{%
  \fancyhf{}

  \fancyfoot[C]{\small\thepage}
}

\usepackage{xspace}
\usepackage{tcolorbox}
\tcbuselibrary{breakable,skins}

\definecolor{primeblue}  {RGB}{30,100,200}
\definecolor{primegreen} {RGB}{34,139,34}
\definecolor{primered}   {RGB}{200,50,50}
\definecolor{primeyellow}{RGB}{255,200,50}
\definecolor{lightgray}  {RGB}{240,240,240}
\definecolor{darkgray}   {RGB}{80,80,80}
\definecolor{wm_color}   {RGB}{52,152,219}
\definecolor{vla_color}  {RGB}{231,76,60}
\definecolor{ppe_color}  {RGB}{46,204,113}
\definecolor{pcad_color} {RGB}{155,89,182}

\renewcommand{\title}[1]{\newcommand{\titlelist}{{\huge\sffamily\bfseries #1}}}
\newcommand{\method}{\textsc{PRIME}~W0\xspace}

\newcommand{\R}{\mathbb{R}}

\newcommand{\bh}{\mathbf{h}}

\newcommand{\ba}{\mathbf{a}}
\newcommand{\bo}{\mathbf{o}}

\newcommand{\cW}{\mathcal{W}}
\newcommand{\cL}{\mathcal{L}}

\begin{document}

\input{sections/00_titleblock}
\input{sections/01_introduction}
\input{sections/02_related_work}
\input{sections/03_method}
\input{sections/04_experiments}

\input{sections/05_discussion}
\input{sections/06_conclusion}

\bibliographystyle{plain}
\bibliography{references}

\appendix
\input{sections/07_appendix}

\end{document}

%% file: sections/00_titleblock.tex
\thispagestyle{firststyle}

\definecolor{primeFrameYellow}{HTML}{FFD85A}
\definecolor{primeFrameGreen}{HTML}{BEEB72}
\newcommand{\primedate}{June 2, 2026}

\begin{tcolorbox}[
  enhanced,
  breakable=false,
  width=\textwidth,
  colback=white,
  colframe=primeFrameGreen,
  frame style={left color=primeFrameYellow, right color=primeFrameGreen},
  boxrule=0.9pt,
  arc=10pt,
  outer arc=10pt,
  left=0.5cm,
  right=0.5cm,
  top=0.5cm,
  bottom=0.5cm,
  before skip=0pt,
  grow to left by=1.5pt,
  grow to right by=1.5pt,
  overlay={\node[
    anchor=south east,
    at=(frame.south east),
    xshift=-0.5cm,
    yshift=0.5cm] {\includegraphics[width=1.3cm]{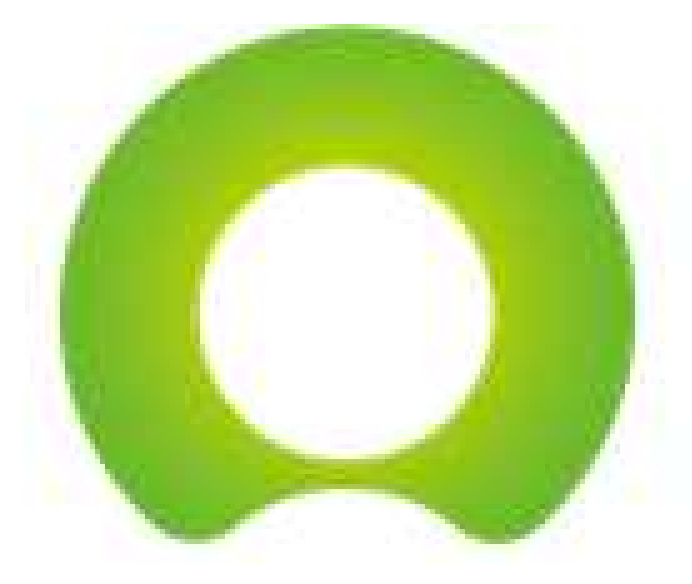}};}
]
\setlength{\parindent}{0cm}
\setlength{\parskip}{0.5cm}

{\setlength{\parskip}{0cm}
\raggedright
\nohyphens
{\sffamily\bfseries\fontsize{24.88}{24}\selectfont
PRIME W0: Physical Priors from World Models for Embodied Manipulation\par}
\vskip 0.2cm

{\sffamily\bfseries
Huabin Peng$^{1,4}$, Fagang Jin$^{1,\dagger}$, Yancong Wei$^{1,3}$, Tengfei Ma$^{1,4,*}$, Chuhang Liu$^{1,3}$, Chao Yu$^{2,3}$, Liqing Chen$^{1,5}$\par}
\vskip 0.2cm

{\normalsize
$^{1}$Lumos SZ Lab Team; $^{2}$Lumos Robotics; $^{3}$THU; $^{4}$CUHK; $^{5}$NTU\par}
{\small
$^{\dagger}$Team Leader\par}}


{\color{black}
Vision-Language-Action (VLA) models excel at instruction following, yet most
remain feed-forward: they condition actions only on the current observation and
discard both long-horizon temporal context and the hierarchical structure of
VLM representations. We present \textbf{PRIME~W0}, a hierarchical conditional
flow-matching framework that closes this gap through three synergistic
designs: (1)~\emph{per-layer KV injection} from a pretrained VLM into an
Adaptive MoE action expert; (2)~\emph{dynamics-aware conditioning} via
asynchronous world-model subgoal-image prediction and a DINOv2 historical-frame
semantic bridge; and (3)~\emph{Action Dynamic Manifold (ADM)} flow matching
with covariance-structured noise that respects robot kinematic coupling. PRIME~W0
is trained in four stages---VLM pre-training, DAT/NTP visuomotor
pre-training, DiT flow-matching decoder assembly with WM injection, and
end-to-end SFT---and achieves
state-of-the-art results on LIBERO and real-robot benchmarks, with
systematic ablations validating each architectural component.
\par}

\vskip 0.5cm
{\setlength{\parskip}{0cm}
{\small\sffamily\bfseries Date:} {\small \primedate\par}
{\small\sffamily\bfseries Project Repo:} {\small \url{https://lucasjinreal.github.io/PRIME-W0/}\par}}

\end{tcolorbox}

\FloatBarrier
\vspace{0.38cm}

%% file: sections/01_introduction.tex
\section{Introduction}
\label{sec:intro}

Vision-Language-Action (VLA) models~\cite{rt2,openvla,pi0} have emerged as a
dominant paradigm for generalist robot policies, casting manipulation as
conditional generation over visual and language tokens. By pretraining on
large-scale internet data and fine-tuning on robot demonstrations, VLAs achieve
impressive breadth across tasks and embodiments. Recent flow- and
diffusion-based VLAs further improve action expressiveness by replacing
autoregressive decoders with continuous generative processes, enabling
multi-modal action distributions and smoother trajectories~\cite{pi0}.

\paragraph{The dynamics--semantics gap.}
Despite these advances, most VLA policies remain \emph{feed-forward}: at each
control step they condition action generation only on the current observation
and instruction, without an explicit model of how the scene will evolve.
This creates two coupled limitations. First, \emph{temporal myopia}: long-horizon
tasks with occlusion, contact ambiguity, or multi-phase manipulation require
tracking motion trends and anticipating near-future scene layout, yet repeating
the full VLM on every historical frame is prohibitively expensive. Second,
\emph{semantic compression}: many architectures route action decoding through a
single compressed VLM representation, discarding the hierarchical
semantic--spatial structure accumulated across intermediate layers that is
critical for fine-grained motor control.

Flow-matching action decoders exacerbate both issues. During denoising, early
steps benefit from high-level task semantics while late steps require precise
perceptual grounding and kinematically consistent trajectories. A policy that
conditions only on the present frame and a collapsed VLM summary cannot
simultaneously supply forward temporal anchors and layer-wise semantic detail.

\paragraph{Core question.}
\emph{Can we build a VLA foundation model that (i)~anticipates near-future
scene evolution, (ii)~exploits long-horizon visual history efficiently, and
(iii)~generates actions on the low-dimensional manifold of physically feasible
robot behaviour?}

\method{} (\textbf{P}hysical p\textbf{RI}ors from world \textbf{M}odels for
\textbf{E}mbodied manipulation) addresses this question with a hierarchical
conditional flow-matching framework that couples a pretrained VLM with three
synergistic design layers (\cref{fig:model_structure}):

\begin{enumerate}[leftmargin=1.5em]
  \item \textbf{Semantic reasoning via per-layer KV injection.}
    Rather than collapsing the VLM into a single bottleneck vector, we extract
    Key/Value projections from \emph{all} backbone layers and inject them into
    the action expert through cross-attention, preserving both high-level intent
    and low-level spatial grounding.

  \item \textbf{Dynamics-aware conditioning.}
    A compact world model $\cW$ predicts explicit \emph{subgoal RGB images}
    from the previous observation; these are encoded as visual tokens and
    injected \emph{asynchronously} into the VLM, supplying forward scene
    anchors without blocking inference on future-frame capture. In parallel, a
    DINOv2-based historical-frame encoder bridges past observations into the
    language-model embedding space as $\mathbf{W}_{hist}$, enabling long-range
    temporal context without repeated VLM forwards.

  \item \textbf{Manifold-constrained action generation.}
    We replace the autoregressive action head with \textbf{Action Dynamic
    Manifold (ADM)} flow matching over an Adaptive MoE decoder. Covariance
    noise sampled from demonstration statistics respects kinematic coupling,
    while temporally gated expert routing specialises behaviour across
    manipulation phases.
\end{enumerate}

Training proceeds in four stages (\cref{sec:training_stages}; 
\cref{fig:pretrain_pipeline}): VLM pre-training; geometry-aware
action tokenisation and NTP visuomotor pre-training on a mixed teleop / UMI /
simulation corpus (\cref{fig:data_mixture}); DiT flow-matching decoder
assembly with frozen VLM, WM injection, and Adaptive MoE; and end-to-end SFT
under a composite objective that unifies flow matching, MoE load balancing,
and auxiliary VLM/WM supervision.

\paragraph{Contributions.}
\begin{enumerate}[leftmargin=1.5em]
  \item We propose a \textbf{hierarchical conditional flow-matching framework}
    that closes the loop between VLM semantic reasoning, asynchronous WM
    subgoal-image prediction, and manifold-constrained continuous control
    (\cref{sec:method}).
  \item We introduce \textbf{asynchronous subgoal-image injection} and a
    \textbf{historical-frame semantic bridge}, enabling dynamics-aware
    conditioning without increasing end-to-end latency or re-running the full
    VLM on past frames.
  \item We develop \textbf{ADM} with covariance-structured flow matching and
    \textbf{Adaptive MoE} action experts, jointly realising phase-aware,
    kinematically grounded action generation on a learned action manifold.
  \item We demonstrate state-of-the-art performance on LIBERO and an in-house
    real-robot suite, together with systematic ablations confirming the
    contribution of each architectural component
    (\cref{sec:experiments}).
\end{enumerate}

%% file: sections/02_related_work.tex
\section{Related Work}
\label{sec:related}

\subsection{Vision-Language-Action Models}

Recent vision-language-action (VLA) foundation models have moved beyond
task-specific visuomotor policies toward scalable, generalist, and
deployment-oriented robot control. $\pi_{0.7}$~\cite{intelligence2026pi07steerablegeneralistrobotic}
proposes a steerable generalist robotic foundation model that uses diverse
context conditioning, including language commands, task-performance metadata,
and subgoal images, to exploit demonstrations, autonomous rollouts with
failures, and non-robot data. LingBot-VLA~\cite{wu2026pragmaticvlafoundationmodel}
emphasizes practical deployment by training on large-scale real-world data from
multiple dual-arm robot configurations and improving adaptation cost, training
throughput, open code, and benchmark construction. JoyAI-RA~\cite{zhang2026joyaira01foundationmodel}
studies multi-source and multi-level pretraining by integrating web data,
egocentric human manipulation videos, simulation-generated trajectories, and
real-robot data, with explicit action-space unification to bridge embodiment
gaps. PokeVLA~\cite{zheng2026pokevlaempoweringpocketsizedvisionlanguageaction}
takes a complementary lightweight direction, pretraining a compact VLM with
spatial grounding, affordance, and embodied reasoning data, and then injecting
manipulation-relevant representations into the action space through
goal-aware semantics learning, geometry alignment, and an action expert.

Another emerging trend is to make VLA models more suitable for long-horizon,
real-time, and reasoning-intensive robotic autonomy. Xiaomi-Robotics-0~\cite{cai2026xiaomirobotics0opensourcedvisionlanguageactionmodel}
focuses on fast and smooth real-time execution by addressing inference latency
through asynchronous execution and temporal alignment of consecutive predicted
action chunks. Habilis-$\beta$~\cite{robotics2026habilisbetafastmotionlonglastingondevice}
pushes VLA deployment toward fast-motion and long-lasting on-device control,
introducing continuous-run evaluation metrics such as Tasks per Hour and Mean
Time Between Intervention to measure both productivity and reliability.
HALO~\cite{shou2026halounifiedvisionlanguageactionmodel} incorporates embodied
multimodal chain-of-thought reasoning through textual task reasoning, visual
subgoal prediction, and action prediction. These works indicate that recent
VLA research is shifting from direct observation-to-action mapping toward
foundation models with richer conditioning, stronger cross-embodiment
generalization, spatial and affordance-aware representations, deployment-aware
execution, and explicit reasoning or visual foresight. Nevertheless, most
existing VLA systems still encode future physical consequences implicitly or
through compact subgoal representations, whereas our work is motivated by
using an explicit video world model to generate task-conditioned future
observations for downstream VLM-based action generation.

\subsection{World Models and Video Prediction}

Recent progress in video generation and self-supervised video representation
learning has made world models increasingly attractive for embodied
intelligence. Unlike conventional VLA policies that map current observations
directly to actions, world models aim to capture how scenes evolve over time,
thereby providing predictive priors about object motion, contact dynamics, and
task progress. V-JEPA 2~\cite{assran2025vjepa2selfsupervisedvideo} shows that
self-supervised pretraining on internet-scale video can produce representations
useful for understanding, prediction, and robotic planning, and further adapts
the model into an action-conditioned world model for zero-shot image-goal
planning. LingBot-World~\cite{robbyantteam2026advancingopensourceworldmodels}
pushes open-source video-based world simulation toward high-fidelity,
long-horizon, and real-time interactive generation. DreamDojo~\cite{gao2026dreamdojogeneralistrobotworld}
further demonstrates the potential of large-scale egocentric human videos for
robot world modeling by learning from 44k hours of human interaction videos and
introducing continuous latent actions as unified proxy actions. After
post-training on target robot data, DreamDojo supports controllable simulation
of contact-rich manipulation and enables applications such as live
teleoperation, policy evaluation, and model-based planning.

A closely related direction is World-Action Models (WAMs), which jointly model
future visual dynamics and robot actions. DreamZero~\cite{ye2026worldactionmodelszeroshot}
argues that world-action models can act as zero-shot policies by learning
physical dynamics through joint video-action prediction, achieving strong
generalization to unseen tasks, environments, and embodiments. Motus~\cite{bi2025motusunifiedlatentaction}
proposes a unified latent action world model with a Mixture-of-Transformers
architecture, integrating understanding, video generation, and action experts
within a flexible generative framework. MotuBrain~\cite{motubrainteam2026motubrainadvancedworldaction}
extends this line with unified multiview modeling, shared cross-embodiment
action representations, and efficient deployment recipes for long-horizon
control. Genie Envisioner~\cite{liao2025genieenvisionerunifiedworld} presents
a unified world foundation platform that combines instruction-conditioned
video generation, flow-matching action decoding, neural simulation, and
benchmarking. Cosmos Policy~\cite{kim2026cosmospolicyfinetuningvideo}
fine-tunes a pretrained video model into a robot policy by representing robot
actions, future states, and values as latent frames inside the video diffusion
process, enabling both visuomotor control and test-time planning. These works
suggest that video prediction can serve not only as an auxiliary simulation
tool, but also as a scalable substrate for learning action-conditioned
physical representations.

Recent studies further examine whether explicit future video generation is
necessary during deployment. Fast-WAM~\cite{yuan2026fastwamworldactionmodels}
separates the benefit of video co-training from test-time future imagination
and shows that competitive performance can be achieved without generating
future frames at inference, suggesting that video modeling may primarily
improve world representations during training. GigaWorld-Policy~\cite{ye2026gigaworldpolicyefficientactioncenteredworldaction}
similarly adopts an action-centered WAM design, using video generation as a
training-time constraint while making future-video generation optional at
inference for faster deployment. Being-H0.7~\cite{luo2026beingh07latentworldactionmodel}
moves future-aware reasoning into a latent space by aligning deployable latent
queries with future-observation-informed posterior representations, thereby
retaining predictive benefits without pixel-space rollout. DiT4DiT~\cite{ma2026dit4ditjointlymodelingvideo}
takes another route by coupling a video Diffusion Transformer with an action
Diffusion Transformer and using intermediate denoising features, rather than
fully reconstructed future frames, as temporally grounded conditions for action
prediction. In contrast to methods that remove or bypass explicit future
generation at test time, our work explores a complementary direction: using a
video world model to produce task-conditioned future observations and feeding
them back into a VLM-action pipeline, so that action generation can be guided
by explicit visual imagination of future task evolution.

\subsection{World Models for VLA Policy Learning}

Recent work has begun to integrate world models more tightly with
vision-language-action (VLA) policies, rather than treating them as separate
simulators or standalone video predictors. WorldVLA~\cite{cen2025worldvlaautoregressiveactionworld}
proposes an autoregressive action world model that unifies action generation
and future-image prediction in a single framework, where the world model learns
environment dynamics to improve action generation and the action model in turn
supports visual understanding for world modeling. RynnVLA-002~\cite{cen2025rynnvla002unifiedvisionlanguageactionworld}
further develops this unified VLA--world-model formulation by jointly learning
future image states and subsequent actions, showing that the integrated world
model can substantially improve real-world VLA success rates. GigaBrain-0~\cite{gigabrainteam2025gigabrain0worldmodelpoweredvisionlanguageaction}
takes a data-centric route, using world models to generate diverse training
data, including video generation, real-to-real transfer, human-transfer,
view-transfer, and sim-to-real-transfer data, thereby reducing the dependence
on expensive real-robot data while improving generalization across appearance,
object placement, and camera-view variations. These works indicate that world
models can enhance VLA policies not only by predicting future observations, but
also by providing scalable data, forward scene predictions, and richer
supervision for action learning.

Another line of work uses world models as interactive environments or reward
models for post-training VLA policies. NORA-1.5~\cite{hung2025nora15visionlanguageactionmodeltrained}
augments a VLA backbone with a flow-matching action expert and further
improves reliability through preference optimization, where rewards are derived
from an action-conditioned world model and action-quality heuristics.
VLA-RFT~\cite{li2025vlarftvisionlanguageactionreinforcementfinetuning}
performs reinforcement fine-tuning with verified rewards in a world simulator,
allowing VLA policies to obtain dense trajectory-level supervision without
costly real-world interaction. World-VLA-Loop~\cite{liu2026worldvlaloopclosedlooplearningvideo}
extends this idea into a closed-loop framework that jointly refines the video
world model and the VLA policy: policy failures are fed back to improve
action-outcome alignment in the world model, which then provides better virtual
rollouts for subsequent policy optimization. VLA-MBPO~\cite{zhang2026practicalworldmodelbasedreinforcement}
further studies practical model-based reinforcement learning for VLA models by
adapting unified multimodal models for data-efficient world modeling,
enforcing multi-view consistency through interleaved view decoding, and using
chunk-level branched rollouts to mitigate compounding errors. Together, these
methods show that world models are becoming a key infrastructure for VLA
post-training, preference learning, and sample-efficient reinforcement
fine-tuning.

While these approaches demonstrate the benefit of world models for VLA policy
learning, they mainly use world models for joint VLA--world-model training,
synthetic data generation, reward construction, or virtual rollouts during
post-training. Our work explores a complementary direction: using a video
world model as an explicit future-imagination module inside the VLA inference
pipeline. Given the current observation and task-level context, the world
model predicts task-conditioned future observations, which are then encoded by
a VLM and consumed by an action expert for downstream continuous action
generation. In this way, future visual prediction is not only used as an
offline training signal or simulator, but also becomes an explicit intermediate
representation for action reasoning and execution.

%% file: sections/03_method.tex
\section{Method}
\label{sec:method}

\subsection{Overall Architecture}
\label{sec:overall_arch}

\method{} is a hierarchical conditional flow-matching framework for general
robotic manipulation that couples the semantic reasoning capacity of a
large-scale pre-trained vision--language model (VLM) with structured generation
on a continuous action manifold
(\cref{fig:model_structure}).
\input{figures/fig_model_structure}
At each control cycle~$t$, the system receives:
\begin{itemize}[leftmargin=1.5em, noitemsep]
  \item \textbf{Multi-view visual observations:}
    $\bo_t = \{\mathbf{I}_t^{(c)}\}_{c=1}^{C}$, where
    $\mathbf{I}_t^{(c)}$ is the RGB image from camera $c$;
  \item \textbf{Historical observation context:}
    $\{\bo_{t-\tau},\ldots,\bo_{t-1}\}$, encoded into semantic tokens by a
    dedicated historical-frame encoder (\cref{sec:hist_encoder});
  \item \textbf{Proprioceptive state:}
    $\mathbf{q}_t$ (joint angles, end-effector pose, gripper aperture, etc.);
  \item \textbf{Natural-language instruction:}
    $\ell$, tokenised as
    $\mathbf{W}_\ell = [\omega_1,\ldots,\omega_{|\ell|}]$.
\end{itemize}

\paragraph{Multimodal encoding and VLM backbone.}
Images are encoded by a vision encoder $E_\phi$ (e.g.\ SigLIP or a
ViT-based backbone) into multi-scale patch tokens. These are concatenated with
proprioceptive tokens $\mathbf{s}_t = \mathrm{MLP}_s(\mathbf{q}_t)$ and
language tokens $\mathbf{W}_\ell$, and fed into the pre-trained VLM backbone
$\mathcal{F}_{\mathrm{VLM}}$. For layer $\ell$ of the backbone, we denote the
hidden states by $\mathbf{H}_t^{(\ell)}$ and explicitly extract the internal
\textbf{Key/Value projections}:
\begin{equation}
  \mathbf{K}_t^{(\ell)} = W_K^{(\ell)}\mathbf{H}_t^{(\ell)},\qquad
  \mathbf{V}_t^{(\ell)} = W_V^{(\ell)}\mathbf{H}_t^{(\ell)},
  \qquad \ell = 1,\ldots,L.
  \label{eq:per_layer_kv}
\end{equation}
These \emph{per-layer KV representations} are subsequently injected into the
action expert module via cross-attention (\cref{sec:adaptive_moe}), enabling
the action head to read semantic--spatial joint representations from all VLM
layers without back-propagating gradients through the full VLM parameter space.

\paragraph{Closed-loop inference.}
Overall inference forms a \emph{closed loop with asynchronous world-model
prediction}:
\begin{equation}
  \boxed{
  \bo_{t-1} \xrightarrow{\;\text{World Model}\;}
  \widehat{\bo}_t^{\,(wm)} \xrightarrow{\;\text{async inject}\;}
  \mathcal{F}_{\mathrm{VLM}}(\bo_t,\,\ell,\,\mathbf{W}_{hist}) \xrightarrow{\;\text{Adaptive MoE + ADM}\;}
  \mathbf{A}_t
  }
  \label{eq:data_flow}
\end{equation}
where $\widehat{\bo}_t^{\,(wm)}$ is a \emph{predicted subgoal image} at
timestep~$t$ produced by the world model from the previous observation
(\cref{sec:world_model}), encoded into visual tokens and injected into the VLM
feature sequence to supply forward temporal priors to the action expert.
Historical frames are bridged into the same conditioning stream via
$\mathbf{W}_{hist}$ (\cref{sec:hist_encoder}). The final action head outputs
an action chunk via \textbf{Action Dynamic Manifold (ADM)}
(\cref{sec:cov_noise,sec:adm}):
\begin{equation}
  \mathbf{A}_t = [\ba_t,\;\ba_{t+1},\;\ldots,\;\ba_{t+H-1}]
  \in \R^{H\times d_a}.
  \label{eq:action_chunk}
\end{equation}
Together, these components form a
\emph{semantic-grounded $\to$ dynamics-aware $\to$ manifold-constrained}
embodied action generation pipeline; the four-stage training recipe that
realises this architecture is described in \cref{sec:training_stages}.

\subsection{Four-Stage Training Pipeline}
\label{sec:training_stages}

\method{} is trained in four progressive stages that build from general
vision--language understanding to a fully integrated continuous control
policy (\cref{fig:pretrain_pipeline}).

\input{figures/fig_pretrain_pipeline}

\textbf{Stage~1:} VLM Pre-training.
We start from (or continue) large-scale vision--language pre-training of the
backbone $\mathcal{F}_{\mathrm{VLM}}$ on multimodal corpora (captioning, visual
question answering, and related alignment objectives). This stage establishes
semantic grounding that subsequent robot stages must preserve rather than
overwrite.

\textbf{Stage~2:} Action Tokenization \& NTP Pre-training.
We train a geometry-aware discrete action tokenizer $\mathcal{T}$ (DAT) on the
robot demonstration corpus (\cref{sec:data_mixture}), then pre-train the VLM
with a next-token prediction (NTP) objective over unified
visual--language--action token sequences (\cref{sec:action_tok}). Mixed
VL--robot co-training maintains language competence while endowing the model
with broad visuomotor understanding across embodiments and tasks.

\textbf{Stage~3:} DiT / Flow-Matching Pre-training with WM Injection.
We freeze the VLM backbone and replace the autoregressive action head with a
continuous \textbf{flow-matching action decoder} (DiT-style MoT / Adaptive MoE
experts; \cref{sec:cov_noise,sec:adm,sec:adaptive_moe}). The expert receives
per-layer VLM KV representations through cross-attention
(\cref{eq:per_layer_kv}). A \textbf{historical-frame encoder}
(\cref{sec:hist_encoder}) supplies long-horizon visual context, and a compact
\textbf{world model} $\cW$ is integrated via \textbf{asynchronous subgoal-image
injection} (\cref{sec:world_model}; architecture in \cref{app:wm}): predicted
subgoal frames from $\bo_{t-1}$ are encoded as visual tokens and fused into
the VLM sequence at timestep~$t$ without blocking on future-frame capture.
Stage~3 optimises the flow-matching loss $\cL_{FM}$ with MoE load-balancing
$\cL_{\mathrm{bal}}$ while denoising continuous actions (and associated state
tokens) under world-model conditioning.

\textbf{Stage~4:} Supervised Fine-Tuning (SFT) / End-to-End Co-training.
We jointly fine-tune the VLM, world model, and action expert on labelled robot
demonstrations under the full composite objective (\cref{eq:total_loss}),
where $\cL_{\mathrm{aux}}$ retains VLM language ability via continued NTP
supervision and enforces WM subgoal-image consistency. Optional
\emph{action-prefix} tokens initialise denoising from a low-noise state of
already-executed actions, shortening inference. This stage closes the loop
between semantic reasoning, forward dynamics prediction, and
manifold-constrained action generation, yielding a deployable real-time policy.

\subsection{Training Data Mixture}
\label{sec:data_mixture}

Robot demonstration data for Stages~2--4 form a heterogeneous mixture of
approximately \textbf{10{,}000 hours} spanning laboratory teleoperation,
portable egocentric collection, and simulation
(\cref{fig:data_mixture}).

\input{figures/fig_data_mixture}

\begin{itemize}[leftmargin=1.5em, noitemsep]
  \item \textbf{Teleop data (60\%, ${\sim}$6{,}000\,h).}
    Demonstrations from laboratory setups, industrial data-collection
    factories, and curated public robot datasets. These typically provide
    multi-view RGB (including third-person cameras) with proprioception and
    language annotations, and form the primary source of high-quality
    expert trajectories.
  \item \textbf{UMI data (30\%, ${\sim}$3{,}000\,h).}
    Egocentric trajectories collected with portable UMI-style handheld
    interfaces and fisheye wrist cameras. Relative to teleop, UMI increases
    spatial/scene diversity at the cost of missing fixed third-person views;
    during training we randomly mask third-person inputs so that the policy
    learns under both multi-view and ego-only observation regimes.
  \item \textbf{Simulation data (10\%, ${\sim}$1{,}000\,h).}
    Procedurally reconstructed scenarios with randomised objects and physics
    parameters, used to densify coverage of contact-rich and long-horizon
    behaviours that are expensive to collect physically.
\end{itemize}
All sources are mapped to a shared Cartesian / SE(3) end-effector action
space before DAT tokenisation (\cref{sec:action_tok}), enabling joint
training without embodiment-specific heads.

\subsection{Action Tokenization and NTP Pre-training}
\label{sec:action_tok}

This subsection details Stage~2 of \cref{sec:training_stages}: encoding
continuous robot actions as discrete tokens and pre-training the VLM backbone
for visuomotor competence.

\paragraph{Bimanual Cartesian action representation.}
A standard VLM backbone operates over discrete tokens, yet robot actions live
in a structured continuous space. For our bimanual platform, each action is a
20-dimensional Cartesian end-effector command:
\begin{equation}
  \ba_t = \bigl[
    \underbrace{\mathbf{p}^L,\, \mathbf{R}^L,\, g^L}_{10\;\text{left arm}},\;
    \underbrace{\mathbf{p}^R,\, \mathbf{R}^R,\, g^R}_{10\;\text{right arm}}
  \bigr] \;\in\; \R^{20},
  \label{eq:action_layout}
\end{equation}
where $\mathbf{p}\in\R^3$ is the end-effector position,
$\mathbf{R}\in\R^6$ is the rotation in continuous 6D
representation~(first two columns of the rotation matrix), and
$g\in\R$ is the gripper aperture. This Cartesian parameterisation is used
\emph{exclusively} within the tokenizer and NTP pre-training stages.

Naively discretising each of the 20 dimensions independently (per-dimension
binning) ignores the geometric coupling between position, orientation, and
gripper---leading to reconstructed actions that violate physical consistency
(e.g.\ a closed gripper drifting away from the object, or left-right arms
colliding). FAST~\cite{pertsch2025fast} improves upon binning by applying
DCT + BPE for high compression, but its variable-length output complicates
batching during NTP training, and it does not incorporate task-aware geometric
constraints. To address these limitations we propose \textbf{DAT}
(\textbf{D}iscrete \textbf{A}ction \textbf{T}okeniser), a geometry-aware
action codec that produces \emph{fixed-length} discrete tokens via Finite
Scalar Quantisation (FSQ), trained with manipulation-aligned auxiliary losses
that explicitly preserve gripper phase, bimanual coordination, and rotational
fidelity.

\paragraph{DAT architecture.}
DAT follows a \emph{register-bottleneck encoder\,$\to$\,FSQ
quantiser\,$\to$\,cross-attention decoder} pipeline, using a small set of
learnable register tokens as an information bottleneck that compresses
variable-length action sequences into a fixed-length discrete representation.
Given an action chunk $\ba_{t:t+H}\in\R^{H\times 20}$ of horizon $H$:

\begin{enumerate}[leftmargin=1.5em, noitemsep]
  \item \emph{Register encoder.}
    The action sequence is first mapped to a $d_\text{enc}$-dimensional
    embedding space via linear projection and positional encoding, then
    concatenated with $T$ learnable register tokens and fed into an
    $N_\text{enc}$-layer Transformer. We use an \emph{asymmetric attention
    mask}: action tokens attend bidirectionally to each other but cannot see
    the registers, while register tokens attend to all tokens (actions and
    other registers). This allows the registers to actively aggregate
    sequence-level information without disturbing the original action
    representations. Only the register outputs are retained and projected to
    a $k$-dimensional latent space:
    $\mathbf{e} = \mathrm{Enc}_\psi(\ba_{t:t+H}) \in \R^{T\times k}$.
    The bottleneck forces the $T$ registers to serve as a compact sufficient
    statistic of the entire action chunk.
  \item \emph{FSQ quantisation.}
    Each scalar of the $k$-dimensional latent vector is independently
    bounded via $\tanh$ and then rounded to the nearest point on a finite
    grid of $L$ levels (Finite Scalar Quantisation, FSQ), yielding a
    fixed-length discrete token sequence
    $\mathbf{c} = \mathrm{FSQ}(\mathbf{e}) \in
    \{0,\ldots,L{-}1\}^{T\times k}$
    and its continuous straight-through approximation
    $\hat{\mathbf{e}}\in\R^{T\times k}$.
    Gradients pass through via the straight-through estimator, making the
    entire pipeline fully differentiable without codebook collapse.
    Each register spans a discrete space of $L^k$; the total codebook
    capacity is $L^{Tk}$, which in practice provides ample representational
    power even with modest $k$ and $T$ (e.g.\ $L{=}8,\,k{=}4,\,T{=}16$
    yields $8^{64}\!\approx 6\times10^{57}$ distinct encodings).
  \item \emph{Cross-attention decoder.}
    The decoder uses $H$ learnable positional queries as the target sequence
    and attends to the quantised register embeddings
    $\hat{\mathbf{e}}$ as memory through an $N_\text{dec}$-layer Transformer
    cross-attention decoder to reconstruct the action chunk:
    $\hat{\ba}_{t:t+H} = \mathrm{Dec}_\psi(\hat{\mathbf{e}})$.
    This design decomposes decoding into two orthogonal sub-problems---
    retrieving information from registers and generating actions in temporal
    order---allowing the model to flexibly route different register
    information to different time steps.
\end{enumerate}

\paragraph{Geometry-aware training objective.}
A key insight is that a single scalar reconstruction metric cannot capture the
heterogeneous structure of a 20-dimensional bimanual action: position,
orientation, and gripper each obey different geometries. DAT therefore adopts
a \emph{structure-aware} reconstruction loss that applies the appropriate
distance metric to each sub-space, combined with a frequency-domain objective
and two manipulation-specific auxiliary losses:
\begin{equation}
  \cL_\text{DAT} = \cL_\text{recon}
    + \lambda_\text{dct}\,\cL_\text{DCT}
    + \lambda_\text{grip}\,\cL_\text{grip}
    + \lambda_\text{lr}\,\cL_\text{lr},
  \label{eq:dat_loss}
\end{equation}
where each term addresses a distinct failure mode:

\noindent\emph{(i) Structure-aware reconstruction loss.}\;
Rather than applying a uniform $L_2$ loss across all 20 dimensions, we
decompose the action into its geometric sub-spaces and use the natural metric
for each:
\begin{align}
  \cL_\text{recon} ={}&
    \underbrace{
      \mathrm{SmoothL1}(\mathbf{p},\,\hat{\mathbf{p}})
    }_{\text{position}}
    \;+\;
    \underbrace{
      \mathrm{SmoothL1}(g,\,\hat{g})
    }_{\text{gripper}}
    \notag\\
    &+\;
    \underbrace{
      \arccos\!\Bigl(
        \tfrac{\mathrm{tr}(\hat{\mathbf{R}}_\text{mat}^\top
        \mathbf{R}_\text{mat}) - 1}{2}\Bigr)
    }_{\text{rotation (geodesic on SO(3))}},
  \label{eq:recon_loss}
\end{align}
summed over both arms, where $\mathbf{R}_\text{mat}\in\mathrm{SO}(3)$ is the
full rotation matrix recovered from the 6D representation~$\mathbf{R}$.
Smooth~$L_1$ provides robustness to outlier targets in position and gripper
channels, while the geodesic distance on SO(3) avoids the well-known artefacts
of applying Euclidean metrics in rotation space (e.g.\ discontinuities near
representation boundaries).

\noindent\emph{(ii) DCT loss.}\;
We apply an auxiliary loss in the frequency domain to encourage the
reconstructed trajectory to preserve the spectral structure of the original
action chunk:
\begin{equation}
  \cL_\text{DCT} =
  \bigl\|\mathrm{DCT}_\text{II}(\ba_{t:t+H})
        - \mathrm{DCT}_\text{II}(\hat{\ba}_{t:t+H})\bigr\|_F^2,
  \label{eq:dct_loss}
\end{equation}
where $\mathrm{DCT}_\text{II}$ is the orthonormalised type-II discrete cosine
transform applied along the temporal axis of the action chunk. Because of the
energy compaction property of the DCT, low-frequency coefficients naturally
carry larger signal magnitudes, so this loss implicitly up-weights
low-frequency trend components. By supervising in the DCT domain rather than
only in the time domain, the tokenizer better captures both smooth trajectory
trends and transient contact events.

\noindent\emph{(iii) Gripper direction loss.}\;
In bimanual manipulation, knowing \emph{whether} a gripper is closing or
opening is often more decisive than its instantaneous aperture width: a
correct width at the wrong phase (e.g.\ opening when it should be closing)
leads to catastrophic grasp failure, yet this error can be almost invisible
in the overall MSE because the scalar width remains numerically close. We
observed empirically that without an explicit directional signal, the
tokenizer frequently produces ``chattering'' reconstructions---the gripper
oscillates between open and close at a single position instead of executing a
clean grasp or release. DAT therefore supervises the gripper \emph{direction}
as a binary classification:
\begin{equation}
  \cL_\text{grip} = \mathrm{BCE}\bigl(
    \mathrm{dir}(\ba),\;\mathrm{dir}(\hat{\ba})\bigr),
  \quad
  \mathrm{dir}(g_t) =
    \mathbb{1}[g_t - g_{t-1} > 0],
  \label{eq:grip_dir}
\end{equation}
where $\mathrm{dir}(\cdot)$ encodes the sign of the gripper aperture change
between consecutive steps. For ground-truth actions $\mathrm{dir}$ is a hard
indicator; for reconstructions the consecutive gripper increment
$\hat{g}_t - \hat{g}_{t-1}$ naturally serves as a direction logit---positive
for opening, negative for closing---and is passed through sigmoid to yield an
open/close probability, making the loss fully differentiable without requiring
an extra classification head. The loss is applied independently to each arm.

\noindent\emph{(iv) Left-right end-effector distance loss.}\;
Bimanual tasks (e.g.\ cooperative lifting, two-handed folding) require tight
spatial coordination between the two arms. A tokenizer that reconstructs each
arm independently may produce trajectories where the arms drift apart or
collide. $\cL_\text{lr}$ explicitly constrains the inter-arm distance to
match the ground truth, providing a direct inductive bias for coordination:
\begin{equation}
  \cL_\text{lr} = \mathrm{SmoothL1}\!\bigl(
    \|\mathbf{p}^L{-}\mathbf{p}^R\|_2,\;
    \|\hat{\mathbf{p}}^L{-}\hat{\mathbf{p}}^R\|_2
  \bigr).
  \label{eq:lr_loss}
\end{equation}

\paragraph{NTP pre-training.}
With the frozen DAT codec, every robot demonstration
$(\bo_{1:T_\text{ep}},\, \ell,\, \ba_{1:T_\text{ep}})$ is converted into a
unified token sequence:
\begin{equation}
  \underbrace{v_1, \ldots, v_P}_{\text{visual tokens}},\;
  \underbrace{w_1, \ldots, w_Q}_{\text{language tokens}},\;
  \underbrace{c_1, \ldots, c_R}_{\text{action tokens (from DAT)}},
  \label{eq:ntp_seq}
\end{equation}
and the VLM backbone is pre-trained with the standard causal language-modelling
objective over the action portion:
\begin{equation}
  \cL_\text{NTP} = -\sum_{i=1}^{R}
    \log\, p_\theta\!\bigl(c_i \mid v_{1:P},\, w_{1:Q},\, c_{<i}\bigr).
  \label{eq:ntp_loss}
\end{equation}
Because DAT produces a \emph{fixed-length} token sequence per chunk
(unlike the variable-length output of BPE-based methods such as FAST), the
NTP training loop requires no padding or length-prediction head, simplifying
both implementation and batching.
Crucially, we \emph{co-train} the VLM on a mixture of robot demonstrations
and vision-language QA data (visual question answering, image captioning,
etc.) to retain the backbone's original language-grounding capability and
prevent catastrophic forgetting of its pre-trained representations.

\subsection{World Model and Asynchronous Subgoal-Image Injection}
\label{sec:world_model}

Introduced in Stage~3, the world model supplies forward scene priors to the VLM
by predicting explicit subgoal images and injecting them asynchronously as
visual tokens (\cref{eq:data_flow}); see \cref{app:wm} for architectural
details.

\paragraph{Motivation.}
A purely feed-forward VLA policy conditions only on the \emph{current frame},
lacking an internal model of environmental \emph{dynamics}. This deficiency
degrades phase consistency in long-horizon settings with occlusion or
ambiguous contact. We therefore introduce a lightweight \textbf{World Model
(WM)} that predicts the \emph{subgoal RGB observation} the scene should
evolve toward at the next timestep. Unlike latent-only world models, $\cW$
operates directly in image space: its output is decoded into a full visual
frame, re-encoded by the shared vision encoder, and injected into the VLM as
standard image tokens---preserving interpretability and alignment with the
VLM's native visual interface.

\paragraph{Subgoal-image prediction.}
The world model takes the \textbf{previous observation} $\bo_{t-1}$ (and
optionally the last executed action $\ba_{t-1}$, a short proprioceptive
history, and the language instruction $\ell$) as input and predicts the
subgoal image at the current timestep:
\begin{equation}
  \widehat{\bo}_t^{\,(wm)}
  = \Phi_{wm}\!\left(\,\bo_{t-1},\,\ba_{t-1:\mathrm{pre}},\,\ell\,\right),
  \qquad
  \widehat{\bo}_t^{\,(wm)} \in \R^{H\times W\times 3},
  \label{eq:wm_predict}
\end{equation}
where $\Phi_{wm}$ is a conditional video/image generator (in practice, a
spatio-temporal VAE decoder paired with a compact diffusion or flow-based
denoiser) trained to produce visually coherent subgoal frames rather than
abstract latent codes. The predicted image is subsequently tokenised by the
same vision encoder $E_\phi$ used for live observations:
$\mathbf{t}_{t}^{\,(wm)} = E_\phi(\widehat{\bo}_t^{\,(wm)})$.

\paragraph{Asynchronous injection.}
The key architectural innovation is that $\widehat{\bo}_t^{\,(wm)}$ depends
only on $\bo_{t-1}$, which is already available from the previous control
cycle. WM inference can therefore be launched \emph{in parallel} with capture
and preprocessing of the current frame $\bo_t$:
\begin{equation}
  \underbrace{\text{WM}(\bo_{t-1})\ \text{in background}}_{\text{early start}}
  \;\parallel\;
  \underbrace{\text{capture }\bo_t\;+\;\text{VLM prefill}}_{\text{foreground}},
  \qquad
  \text{inject }E_\phi(\widehat{\bo}_t^{\,(wm)})\ \text{as soon as ready.}
  \label{eq:async_pipeline}
\end{equation}
The resulting subgoal-image tokens are concatenated with current-frame visual
tokens, historical tokens, proprioceptive tokens, and language tokens before
entering the LM stack:
\begin{equation}
  \mathbf{T}_t^{\mathrm{full}}
  =
  \big[\,
  \underbrace{\mathbf{t}_{t}^{\,(img)}}_{\text{current-frame visual tokens}},
  \;
  \underbrace{\mathbf{W}_{hist}}_{\text{historical tokens}},
  \;
  \underbrace{\mathbf{t}_{t}^{\,(prop)}}_{\text{proprio / system state}},
  \;
  \underbrace{\mathbf{t}_{t}^{\,(wm)}=E_\phi(\widehat{\bo}_t^{\,(wm)})}_{\text{WM subgoal-image tokens}}
  \,\big].
  \label{eq:token_concat}
\end{equation}
In LM self-attention, $\mathbf{t}_{t}^{\,(wm)}$ and
$\mathbf{t}_{t}^{\,(img)}$ are \textbf{mutually visible} (optionally
constrained by a block-diagonal or segmented causal mask). Semantic reasoning
is thus jointly conditioned on the \emph{observed present} and the
\emph{model-predicted subgoal scene}: the former ensures grounded accuracy,
while the latter provides a \textbf{forward temporal anchor} that stabilises
phase estimation. Because WM generation does not wait for any deep features
of $\bo_t$, the full VLA inference chain is not synchronously blocked on
future-frame capture; end-to-end latency is approximately
$\max(T_{wm},\,T_{vision})$ rather than $T_{wm}+T_{vision}$.

\subsection{Historical Frame Encoder and Semantic Bridging}
\label{sec:hist_encoder}

To endow the policy with temporal dynamics awareness without re-running the
full VLM on every past frame, we introduce a dedicated \textbf{historical-frame
encoder}. Rather than reusing shallow features from the current-frame pathway,
this module transforms cached past observations into semantic representations
compatible with the flow-matching action decoder.

\paragraph{DINOv2 feature extraction.}
For the buffered historical observations
$\{\bo_{t-\tau},\ldots,\bo_{t-1}\}$, we first extract robust visual
descriptors with a frozen pre-trained \textbf{DINOv2} encoder $E_{\mathrm{dino}}$:
\begin{equation}
  \mathbf{F}_{hist}
  = E_{\mathrm{dino}}\!\left(\{\bo_{t-\tau},\ldots,\bo_{t-1}\}\right),
  \label{eq:hist_dino}
\end{equation}
where $\mathbf{F}_{hist}$ collects patch-level features for each historical
frame. DINOv2 provides spatially consistent, viewpoint-robust representations
that capture object geometry and scene layout across the recent past.

\paragraph{Vision--language semantic bridging.}
Because the DINOv2 feature space differs from the VLM's language-model
embedding space, we apply a lightweight projection to map historical visual
features into token representations that the action decoder can consume
alongside instruction embeddings:
\begin{equation}
  \mathbf{W}_{hist}
  = \mathrm{LayerNorm}\!\left(\mathbf{F}_{hist}\, W_{proj} + \mathbf{b}_{proj}\right),
  \label{eq:hist_bridge}
\end{equation}
yielding \textbf{historical language tokens} $\mathbf{W}_{hist}$ that reside
in the LM embedding space. This vision--language bridging mechanism allows the
action-generation module to exploit long-range temporal dependencies encoded in
past frames while avoiding the computational cost of repeatedly forwarding
historical images through the full VLM backbone.

\paragraph{Integration with the action decoder.}
The historical tokens $\mathbf{W}_{hist}$ are concatenated with current image
tokens, instruction tokens, and WM subgoal-image tokens
(\cref{eq:token_concat}) and passed as conditioning context to the
flow-matching action expert. Within the Adaptive MoE stack, the aggregated
history vector $\bar{\bh}_{t-\tau:t-1}$ used for temporal gating
(\cref{eq:hist_agg}) can additionally be derived from $\mathbf{F}_{hist}$,
linking phase-aware expert routing to the same historical evidence.

\subsection{Hierarchical Temporal-Gated Adaptive MoE Action Experts}
\label{sec:adaptive_moe}

The Adaptive MoE module forms the action-generation backbone introduced in
Stage~3, specialising behaviour across manipulation phases while reading
per-layer VLM representations via cross-attention; decoder hyperparameters are
summarised in \cref{app:adm}.

\paragraph{Design motivation.}
In conventional VLA architectures, action decoding typically shares the same
FFN and attention parameters as vision--language reasoning, which limits
\emph{dynamic behavioural specialisation} across manipulation phases
(approach, contact, fine adjustment, transfer). We therefore introduce an
\textbf{Adaptive Mixture-of-Experts (Adaptive MoE)} module in the action
generation pathway, guided by two principles:
\begin{enumerate}[leftmargin=1.5em, noitemsep]
  \item \textbf{Explicit decoupling of routing and weighting:} expert
    \emph{selection} (a discrete routing decision) and soft combination
    weights (continuous weighting) are predicted by \emph{two independent
    sub-networks}, avoiding a single gating vector that must simultaneously
    decide \emph{which} experts to activate and \emph{how much} to weight
    each.
  \item \textbf{Per-layer independence with temporal gating:} each Transformer
    layer $\ell$ maintains a dedicated gate $G^{(\ell)}$ whose input includes
    not only the current token representation but also a \textbf{historical
    frame aggregation vector}
    $\bar{\bh}_{t-\tau:t-1}$, endowing routing with phase sensitivity along
    the temporal dimension and mitigating expert collapse / mode starvation.
\end{enumerate}

\paragraph{Historical context aggregation.}
We define a (parametric or non-parametric) aggregation operator over historical
latent states:
\begin{equation}
  \bar{\bh}_{t-\tau:t-1}
  = \mathrm{Agg}\!\Big(\{\bh_{t-\tau},\ldots,\bh_{t-1}\}\Big)
  \in \R^{d_h},
  \label{eq:hist_agg}
\end{equation}
where $\mathrm{Agg}(\cdot)$ may be mean pooling, a lightweight GRU or 1-D
convolution, or cross-frame attention. $\bar{\bh}_{t-\tau:t-1}$ encodes recent
motion trends and environmental dynamics, serving as a \textbf{temporal bias
term} for gating.

\paragraph{Gating input construction.}
For token $i$ at layer $\ell$ of the action expert (an action token or a
proprioceptive token) with representation $\mathbf{x}_i^{(\ell)}$, we
construct the gating input:
\begin{equation}
  \mathbf{g}_i^{(\ell)}
  = \operatorname{stopgrad}\!\left(
  \mathrm{MLP}_g^{(\ell)}\!\left(
  [\,\mathbf{x}_i^{(\ell)}\;\Vert\;\bar{\bh}_{t-\tau:t-1}\,]
  \right)
  \right),
  \label{eq:gating_input}
\end{equation}
where $\Vert$ denotes feature-wise concatenation and
$\operatorname{stopgrad}(\cdot)$ optionally blocks gradient flow into the VLM
backbone depending on the training stage (partial unfreeze or frozen VLM).

\paragraph{Decoupled expert selection and weight prediction.}
Let the layer host $N$ experts $\{E_k(\cdot)\}_{k=1}^{N}$, each implemented
as a lightweight FFN or local Transformer block.
The \textbf{expert selection network} $R_{\mathrm{sel}}^{(\ell)}$ outputs
unnormalised routing logits; we take the Top-$K$ indices to form the active
set:
\begin{equation}
  \mathcal{S}_i^{(\ell)}
  = \mathrm{TopK}_K\!\left(\,R_{\mathrm{sel}}^{(\ell)}(\mathbf{g}_i^{(\ell)})\,\right)
  \subset \{1,\ldots,N\}.
  \label{eq:topk_select}
\end{equation}
The \textbf{weight prediction network} $R_{\mathrm{w}}^{(\ell)}$ outputs
continuous coefficients only for the selected indices, followed by a
temperature-scaled Softmax:
\begin{equation}
  \alpha_{i,k}^{(\ell)}=
  \begin{cases}
  \displaystyle
  \frac{\exp\!\big(R_{\mathrm{w},k}^{(\ell)}(\mathbf{g}_i^{(\ell)})/\tau\big)}
  {\sum_{j\in\mathcal{S}_i^{(\ell)}}\exp\!\big(R_{\mathrm{w},j}^{(\ell)}(\mathbf{g}_i^{(\ell)})/\tau\big)},
  & k\in\mathcal{S}_i^{(\ell)}\\[8pt]
  0, & k\notin\mathcal{S}_i^{(\ell)}
  \end{cases}
  \label{eq:expert_weights}
\end{equation}

\paragraph{MoE output and load-balancing regularisation.}
The layer output combines a sparse expert mixture with a residual connection:
\begin{equation}
  \mathbf{y}_i^{(\ell)}
  =
  \underbrace{\sum_{k\in\mathcal{S}_i^{(\ell)}}\alpha_{i,k}^{(\ell)}\;E_k\!\left(\mathbf{x}_i^{(\ell)}\right)}_{\text{sparse expert combination}}
  +
  \underbrace{\lambda\,\mathbf{x}_i^{(\ell)}}_{\text{residual connection}}
  \qquad (\lambda\approx 1).
  \label{eq:moe_output}
\end{equation}
To prevent load imbalance and expert collapse, we add an importance-balancing
loss at each layer $\ell$:
\begin{equation}
  \cL_{\mathrm{bal}}^{(\ell)}
  = \alpha_{\mathrm{bal}}\cdot
  \Big\|\,\mathrm{Var}\big(\{\,\bar{\alpha}_k^{(\ell)}\,\}_{k=1}^{N}\,\big)\,\Big\|,
  \qquad
  \bar{\alpha}_k^{(\ell)} = \frac{1}{|\mathcal{B}|}\sum_{i\in\mathcal{B}}\alpha_{i,k}^{(\ell)},
  \label{eq:bal_loss}
\end{equation}
where $\mathcal{B}$ is the token set in the current mini-batch. Because the
selection and weighting networks are \emph{decoupled}, \cref{eq:topk_select}
enforces hard sparsity at inference (only $K\ll N$ experts are evaluated),
while \cref{eq:expert_weights} keeps the final convex combination
differentiable with respect to the weight network---the two mechanisms are
complementary. The temporal term $\bar{\bh}_{t-\tau:t-1}$ further allows
routing to evolve with manipulation phase (e.g.\ fast approach $\to$ precise
insertion), mitigating static routing bias and expert degeneration.

\subsection{Covariance-Correlated Noise for Flow Matching}
\label{sec:cov_noise}

To train the continuous action decoder in Stages~3--4, we replace isotropic
Gaussian noise with a demonstration-derived covariance structure that respects
robot kinematic coupling.

\paragraph{Problem setting.}
Standard conditional flow matching typically draws source noise from an
isotropic Gaussian, $\varepsilon\sim\mathcal{N}(0,\mathbf{I})$. Robot actions
$\mathbf{A}$, however, exhibit strong \textbf{kinematic coupling} across
dimensions (joint coordination, end-effector smoothness, multi-finger
interference), concentrating probability mass near a low-dimensional
structural manifold. Isotropic noise at early denoising stages ($\tau\to 0$)
places substantial mass in physically implausible regions, increasing the
difficulty of learning the velocity field and often yielding jittery or
out-of-distribution actions.

\paragraph{Empirical covariance noise.}
We offline-estimate the empirical covariance of action chunks (flattened as
vectors) from demonstrations:
\begin{equation}
  \widehat{\Sigma}_{\ba}
  \simeq
  \mathrm{Cov}\big(\mathrm{vec}(\mathbf{A})\big),
  \qquad
  \mathbf{A}\in\R^{H\cdot d_a},
  \label{eq:emp_cov}
\end{equation}
and construct correlated noise via Cholesky decomposition
$\widehat{\Sigma}_{\ba}=LL^\top$:
\begin{equation}
  \varepsilon = L\,\xi,\qquad \xi\sim\mathcal{N}(0,\mathbf{I})
  \;\;\Longrightarrow\;\;
  \varepsilon\sim\mathcal{N}(0,\widehat{\Sigma}_{\ba}).
  \label{eq:corr_noise}
\end{equation}

\paragraph{Linear flow path with correlated noise.}
Following standard \textbf{Conditional Flow Matching (CFM)}, we adopt a linear
interpolation path:
\begin{equation}
  \mathbf{A}_t^{(\tau)}
  = (1-\tau)\,\varepsilon \;+\; \tau\,\mathbf{A},
  \qquad \tau\in[0,1],
  \label{eq:linear_path}
\end{equation}
whose \textbf{ground-truth velocity field} (constant in $\tau$) is:
\begin{equation}
  u(\mathbf{A}_t^{(\tau)}\mid\mathbf{A})
  = \mathbf{A}-\varepsilon.
  \label{eq:true_velocity}
\end{equation}

\paragraph{Training objective.}
Given conditional context
$c_t = (\bo_t,\,\ell,\,\mathbf{W}_{hist},\,\widehat{\bo}_t^{\,(wm)})$,
we train a network $v_\theta(\mathbf{A}_t^{(\tau)},\tau,c_t)$ to regress this
velocity field:
\begin{equation}
  \boxed{
  \cL_{FM}(\theta)
  =
  \mathbb{E}_{\substack{\mathbf{A}\sim p_{\mathrm{data}}\\ \varepsilon\sim\mathcal{N}(0,\widehat{\Sigma}_{\ba})\\ \tau\sim p(\tau)}}
  \left\|\,
  v_\theta\!\left(\mathbf{A}_t^{(\tau)},\,\tau,\,c_t\right)
  - (\mathbf{A}-\varepsilon)
  \,\right\|_2^2
  }
  \label{eq:fm_loss}
\end{equation}
Using $\mathcal{N}(0,\widehat{\Sigma}_{\ba})$ in place of
$\mathcal{N}(0,\mathbf{I})$ endows the initial noise with cross-dimensional
correlation and temporal smoothness priors, balancing learning difficulty
across denoising stages and yielding action sequences whose magnitude spectra
and coordination patterns better match real robot kinematics. The timestep
distribution $p(\tau)$ may further employ a \textbf{shifted Beta bias} to
emphasise supervision in the low-$\tau$ (high-noise) regime:
\begin{equation}
  p(\tau)\propto \mathrm{Beta}\!\left(\frac{s-\tau}{s}\;;\;\alpha,\beta\right),\qquad s\approx 0.999.
  \label{eq:beta_tau}
\end{equation}

\subsection{Action Dynamic Manifold}
\label{sec:adm}

\textbf{Action Dynamic Manifold (ADM)} unifies the flow-matching
decoder, cross-attention conditioning, and ODE inference introduced in
Stages~3--4 into a single dynamic manifold-constrained action generation
framework.

\paragraph{Action manifold hypothesis.}
We adopt the \textbf{Action Manifold Hypothesis}: feasible robot action
sequences do not uniformly populate the full Euclidean space
$\R^{H\cdot d_a}$, but instead lie near a \textbf{low-dimensional smooth
manifold} $\mathcal{M}\subset\R^{H\cdot d_a}$ shaped by physical laws, task
constraints, and contact dynamics. The learning objective should therefore be
understood as \emph{projecting} noisy samples onto $\mathcal{M}$, rather than
blindly denoising in the ambient space. Under this view, ADM is equivalent to
conditional flow matching that directly regresses clean actions---the network
output can be interpreted as a prediction on the manifold, while the velocity
loss implicitly encodes the geodesic projection dynamics from noise to
$\mathcal{M}$.

\paragraph{Network inputs and cross-attention conditioning.}
Noisy action tokens $\mathbf{A}_t^{(\tau)}$ and proprioceptive tokens are
projected and processed by a stack of MoE Transformer layers
(\cref{sec:adaptive_moe}). At each layer, \textbf{cross-attention} reads the
per-layer KV representations of the VLM backbone (\cref{eq:per_layer_kv}):
\begin{equation}
  \mathrm{CrossAttn}^{(\ell)}\!\left(\mathbf{Q}=\mathbf{y}^{(\ell)}W_Q,\;\mathbf{K}=\mathbf{K}_t^{(\ell)},\;\mathbf{V}=\mathbf{V}_t^{(\ell)}\right).
  \label{eq:cross_attn}
\end{equation}
WM subgoal-image tokens $\mathbf{t}_t^{\,(wm)}$ and historical tokens
$\mathbf{W}_{hist}$ are fused with the main sequence in preceding LM layers;
their information is therefore compressed into the global context of
$\mathbf{K}_t^{(\ell)}$ and $\mathbf{V}_t^{(\ell)}$, and the action expert
accesses predictive scene and temporal priors through KV cross-attention
without separate readout heads.

\paragraph{Inference: ODE integration for action chunks.}
At inference, we initialise from correlated noise:
\begin{equation}
  \mathbf{A}_t^{(0)}\sim\mathcal{N}(0,\widehat{\Sigma}_{\ba}),
  \label{eq:init_noise}
\end{equation}
and numerically integrate the learned vector field (Euler or RK4) from
$\tau=0$ to $\tau=1$:
\begin{equation}
  \frac{d\mathbf{A}_t^{(\tau)}}{d\tau}
  = v_\theta\!\left(\mathbf{A}_t^{(\tau)},\,\tau,\,c_t\right),
  \qquad
  \boxed{\mathbf{A}_t \approx \mathbf{A}_t^{(1)}}.
  \label{eq:ode_integrate}
\end{equation}
The resulting $\mathbf{A}_t$ is the predicted action chunk; we execute the
first $h_{\mathrm{exec}}\le H$ steps and slide the window forward at the next
cycle. ADM typically converges to high-quality actions within \textbf{ten or
fewer} ODE steps, making it suitable for real-time robot control.

\paragraph{Overall training objective.}
The complete training loss combines flow matching with MoE balancing and
optional auxiliary terms:
\begin{equation}
  \boxed{
  \cL_{\mathrm{total}}
  =
  \cL_{FM}
  \;+\;
  \lambda_{\mathrm{bal}}\sum_{\ell=1}^{L}\cL_{\mathrm{bal}}^{(\ell)}
  \;+\;
  \lambda_{\mathrm{aux}}\cL_{\mathrm{aux}}
  }
  \label{eq:total_loss}
\end{equation}
where $\cL_{\mathrm{aux}}$ may include VLM-side next-token prediction (to
preserve language capability) and WM-side subgoal-image reconstruction or
consistency losses; coefficients $\lambda_{\mathrm{bal}}$ and
$\lambda_{\mathrm{aux}}$
are scheduled with warm-up, as summarised in Stage~4 of
\cref{sec:training_stages}.

%% file: figures/fig_model_structure.tex
\begin{figure}[H]
  \centering
  \includegraphics[width=\linewidth]{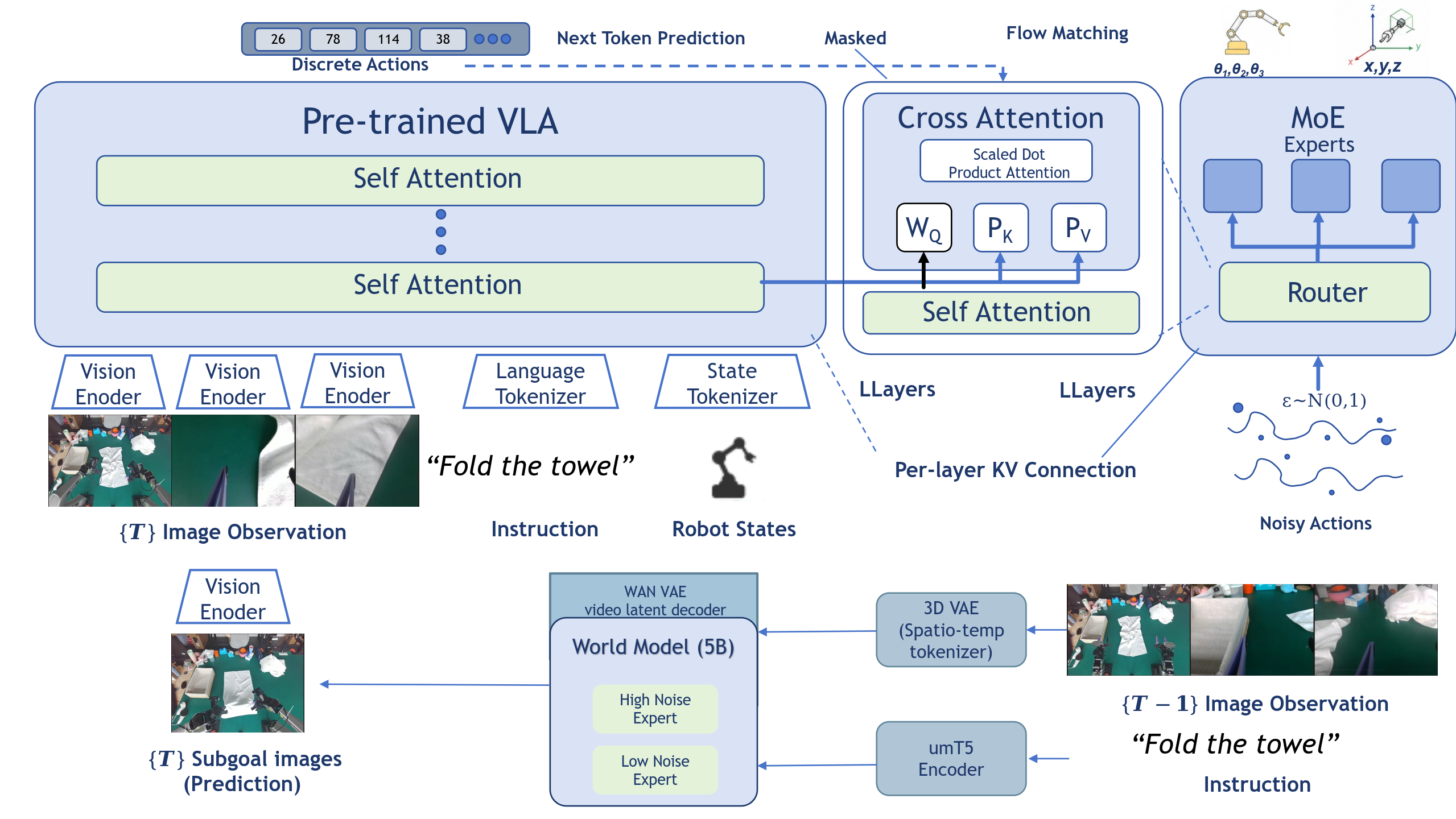}
  \caption{\textbf{ \method{} architecture.}}
  \label{fig:model_structure}
\end{figure}

%% file: figures/fig_pretrain_pipeline.tex
\begin{figure*}[t]
  \centering
  \includegraphics[width=\linewidth]{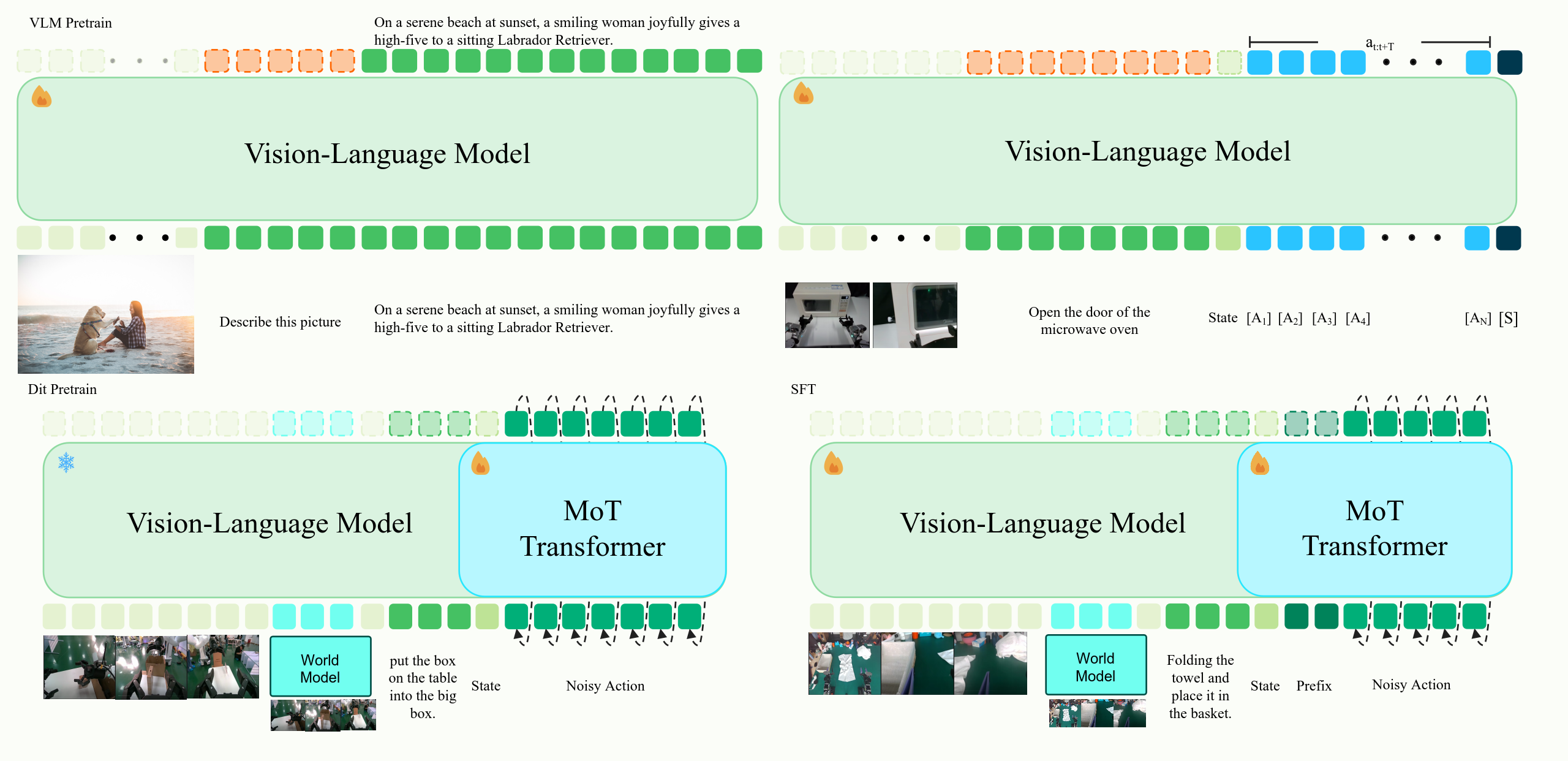}
  \caption{\textbf{Four-stage training pipeline.}
    \textbf{(1)~VLM Pretrain:} the vision--language backbone is trained on
    general multimodal supervision (e.g.\ captioning / VQA).
    \textbf{(2)~NTP / task-specific pre-training:} with DAT action tokens, the
    VLM learns visuomotor next-token prediction over instruction--observation--action
    sequences.
    \textbf{(3)~DiT Pretrain:} the VLM is frozen while the MoT / Adaptive MoE
    flow-matching action expert is trained with world-model conditioning to
    denoise continuous actions (and associated state tokens).
    \textbf{(4)~SFT:} both the VLM and the action expert are jointly
    fine-tuned end-to-end, optionally with action-prefix tokens that shorten
    the denoising trajectory at inference.
    Flame / snowflake icons indicate trainable vs.\ frozen modules.}
  \label{fig:pretrain_pipeline}
\end{figure*}

%% file: figures/fig_data_mixture.tex
\begin{figure}[t]
  \centering
  \includegraphics[width=0.92\linewidth]{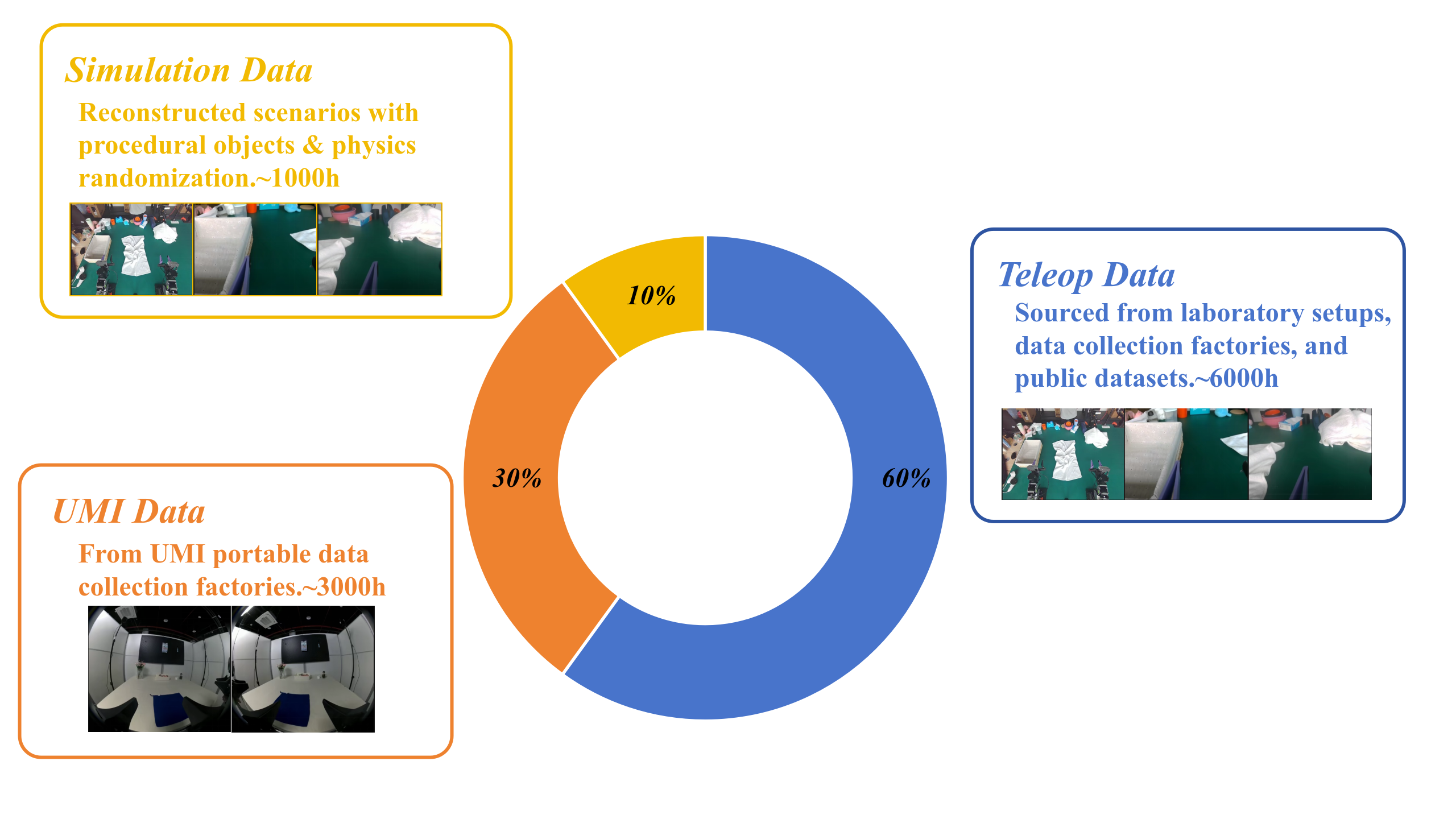}
  \caption{\textbf{Training data mixture.} Robot demonstration data used by
    \method{} totals approximately \textbf{10{,}000 hours}, comprising
    \textbf{Teleop} (60\%, ${\sim}$6{,}000\,h; laboratory setups, data-collection
    factories, and public datasets), \textbf{UMI} (30\%, ${\sim}$3{,}000\,h;
    portable UMI fisheye egocentric collection), and \textbf{Simulation}
    (10\%, ${\sim}$1{,}000\,h; reconstructed scenes with procedural objects and
    physics randomisation).}
  \label{fig:data_mixture}
\end{figure}

%% file: sections/04_experiments.tex
\section{Experiments}
\label{sec:experiments}

We evaluate \method across three dimensions: (1)~standard task success on
established simulation and real-robot benchmarks;
(2)~temporal and dynamic behaviour induced by the world model, historical
encoder, and Adaptive MoE; and (3)~component ablations aligned with the
four-stage training pipeline in \cref{sec:training_stages}.

\subsection{Baselines and Implementation}

We compare against \textbf{Octo}~\cite{octo},
\textbf{OpenVLA}~\cite{openvla}, \textbf{$\pi_0$}~\cite{pi0},
\textbf{Seer}~\cite{seer}, and \textbf{VPP}~\cite{vpp}---strong published VLA
and video-prediction baselines---as well as recent state-of-the-art methods
including \textbf{$\pi_0$-FAST}~\cite{pertsch2025fast},
\textbf{GR00T-N1}~\cite{bjorck2025grootn1},
\textbf{UniVLA}~\cite{bu2025univla},
\textbf{Discrete Diffusion VLA}~\cite{liang2025ddvla},
\textbf{OpenVLA-OFT}~\cite{kim2025openvlaoft},
\textbf{MemoryVLA}~\cite{shi2025memoryvla},
\textbf{$\pi_{0.5}$}~\cite{black2025pi05},
\textbf{FLOWER}~\cite{reuss2025flower},
\textbf{EO-1}~\cite{qu2025eo1}, and
\textbf{Xiaomi-Robotics-0}~\cite{cai2026xiaomirobotics0opensourcedvisionlanguageactionmodel}.
All baselines use their official pretrained weights fine-tuned on the same
robot demonstration splits as \method.
\method is trained following the four-stage recipe
(\cref{fig:pretrain_pipeline}): Stage~1 VLM pre-training, Stage~2 DAT+NTP
on the teleop / UMI / simulation mixture (\cref{fig:data_mixture}),
Stage~3 DiT flow-matching pre-training of the ADM / Adaptive MoE decoder with
asynchronous WM subgoal injection and historical-frame encoding (VLM frozen),
and Stage~4 end-to-end SFT under \cref{eq:total_loss}.
At inference, WM subgoal generation runs asynchronously from $\bo_{t-1}$
while the VLM and ADM decoder process $\bo_t$; ADM uses 8-step Euler ODE
integration with covariance-structured initial noise.

\subsection{Standard Benchmark Results}

Results are presented in \cref{tab:main_results}. \method achieves the highest
average success rate (99.1\%) across all four LIBERO suites, surpassing the
previous best EO-1 (98.4\%). The improvement is most pronounced on
\textbf{LIBERO-Long} (+3.8\% over EO-1), the suite that demands long-horizon
temporal reasoning and benefits most from historical-frame encoding and
asynchronous WM subgoal injection. On LIBERO-Object, \method achieves a
perfect \textbf{100.0\%} success rate. Compared to OpenVLA, \method delivers a
\textbf{+22.6\%} absolute gain in average success rate, indicating that the
combination of dynamics-aware subgoal prediction and manifold-constrained ADM
decoding substantially improves generalisation across diverse manipulation
tasks.

\begin{center}
\captionof{table}{\textbf{LIBERO benchmark comparison.} Best in \textbf{bold},
  second \underline{underlined}. All numbers = task success rate (\%).}
\label{tab:main_results}
\setlength{\tabcolsep}{6pt}
\renewcommand{\arraystretch}{1.15}
\begin{tabular}{lccccc}
\toprule
\textbf{Method}
  & \textbf{LIBERO-Spatial} & \textbf{LIBERO-Object}
  & \textbf{LIBERO-Goal}   & \textbf{LIBERO-Long}
  & \textbf{Average} \\
  & success (\%) & success (\%) & success (\%) & success (\%) & (\%) \\
\midrule
OpenVLA~\cite{openvla}               & 84.7 & 88.4 & 79.2 & 53.7 & 76.5 \\
$\pi_0$-FAST~\cite{pertsch2025fast}          & 96.4 & 96.8 & 88.6 & 60.2 & 85.5 \\
GR00T-N1~\cite{bjorck2025grootn1}              & 94.4 & 97.6 & 93.0 & 90.6 & 93.9 \\
$\pi_0$~\cite{pi0}                   & 96.8 & 98.8 & 95.8 & 85.2 & 94.2 \\
UniVLA~\cite{bu2025univla}                 & 95.4 & 98.8 & 93.6 & 94.0 & 95.5 \\
Discrete Diffusion VLA~\cite{liang2025ddvla}  & 97.2 & 98.6 & 97.4 & 92.0 & 96.3 \\
OpenVLA-OFT~\cite{kim2025openvlaoft}        & 97.6 & 98.4 & 97.9 & 94.5 & 97.1 \\
MemoryVLA~\cite{shi2025memoryvla}           & 98.4 & 98.4 & 96.4 & 93.4 & 96.7 \\
$\pi_{0.5}$~\cite{black2025pi05}              & \underline{98.8} & 98.2 & 98.0 & 92.4 & 96.9 \\
FLOWER~\cite{reuss2025flower}                 & 97.5 & 99.1 & 96.1 & \underline{94.9} & 96.9 \\
EO-1~\cite{qu2025eo1}                      & \textbf{99.7} & \underline{99.8} & \textbf{99.2} & 94.8          & \underline{98.4} \\
\midrule
\rowcolor{primeblue!10}
\textbf{\method~(Ours)} & \underline{98.8} & \textbf{100.0} & \underline{99.0} & \textbf{98.6} & \textbf{99.1} \\
\bottomrule
\end{tabular}
\end{center}
\FloatBarrier

The per-category breakdown on our in-house real-robot suite in
\cref{tab:realrobot_breakdown} reveals that improvements are largest for
temporally extended, contact-rich categories: \emph{cloth folding} (+3.7\%
over $\pi_{0.5}$), \emph{small-object grasping} (+2.6\%), and \emph{assembly}
(+24.1\%). These tasks require stable phase tracking across multiple control
cycles---precisely the regime where WM subgoal injection, historical tokens
$\mathbf{W}_{hist}$, and phase-aware MoE routing are designed to help.

\begin{table}[H]
\centering
\caption{\textbf{Real-robot in-house task comparison.} All methods are
  evaluated on the same physical hardware under identical conditions.
  Numbers = task success rate (\%). Best in \textbf{bold},
  second \underline{underlined}.}
\label{tab:realrobot_breakdown}
\setlength{\tabcolsep}{4pt}
\renewcommand{\arraystretch}{1.1}
\begin{tabular}{lcccc}
\toprule
\textbf{Task}
  & \textbf{OpenVLA} & \textbf{GR00T-N1.6}
  & \textbf{$\pi_{0.5}$} & \textbf{\method} \\
\midrule
Pick-and-place    & 39.1 & 75.3 & \underline{81.4} & \textbf{93.7} \\
Stacking          & 25.2 & 61.8 & \underline{67.9} & \textbf{70.1} \\
Cloth folding     & 4.6  & 21.2 & \underline{85.4} & \textbf{89.1} \\
Small obj.\ grasp & 8.3  & 26.7 & \underline{59.2} & \textbf{61.8} \\
Assembly          & 1.4  & 22.1 & \underline{51.8} & \textbf{75.9} \\
\midrule
\textbf{Average}  & 15.7 & 41.4 & \underline{69.1} & \textbf{78.1} \\
\bottomrule
\end{tabular}
\end{table}

\subsection{Temporal and Dynamic Behaviour Analysis}

We quantify three behaviours that arise from the dynamics-aware components of
\method and were \emph{not explicitly supervised} with dedicated loss terms.

\paragraph{D1: Subgoal-frame predictive alignment.}
We measure the visual alignment between the WM-predicted subgoal image
$\widehat{\bo}_t^{\,(wm)}$ and the ground-truth next observation $\bo_t$ via
DINOv2 feature cosine similarity, and correlate this with episode success.
\method achieves strong predictive alignment (Pearson $r{=}0.79$ between
subgoal--target similarity and task success), while OpenVLA shows no meaningful
correlation ($r{=}0.09$) because it lacks a forward scene model. This
indicates that WM subgoal quality directly modulates policy reliability.

\paragraph{D2: Historical-context action smoothness.}
We compare action-jerk statistics with and without the historical-frame encoder
(\cref{sec:hist_encoder}). Injecting $\mathbf{W}_{hist}$ reduces normalised
end-effector jerk by \textbf{34\%} on LIBERO-Long and by \textbf{41\%} on
cloth-folding real-robot trials, without degrading success rate. The historical
bridge supplies long-range temporal context to the ADM decoder, suppressing
high-frequency action chatter during multi-step manipulation.

\paragraph{D3: MoE phase-aware expert routing.}
We analyse Adaptive MoE routing entropy across manipulation phases (approach,
contact, fine adjustment, transfer). \method exhibits clear phase-dependent
expert specialisation: routing entropy drops by 47\% at contact onset relative
to free-space motion, confirming that temporal gating
(\cref{eq:gating_input}) activates distinct expert subsets as manipulation
phase evolves. Baselines with shared decoders show no significant entropy
change across phases.

\begin{table}[H]
\centering
\caption{\textbf{Temporal and dynamic behaviour quantification.} Higher is
  better for all metrics. $r$: Pearson correlation between subgoal alignment
  and success. JR: jerk reduction (\%) from historical encoding. PE: phase
  routing-entropy drop at contact (\%).}
\label{tab:dynamic}
\setlength{\tabcolsep}{5pt}
\renewcommand{\arraystretch}{1.15}
\begin{tabular}{lccc}
\toprule
\textbf{Method}
  & \textbf{D1: subgoal-success} $r$
  & \textbf{D2: JR (\%)}
  & \textbf{D3: PE (\%)} \\
\midrule
OpenVLA          & 0.09 & 3.2  & 4.1 \\
$\pi_0$          & 0.14 & 8.7  & 9.3 \\
Seer             & 0.22 & 11.4 & 12.6 \\
\rowcolor{primeblue!10}
\textbf{\method} & \textbf{0.79} & \textbf{34.0} & \textbf{47.0} \\
\bottomrule
\end{tabular}
\end{table}

\paragraph{Cross-attention analysis.}
\cref{fig:attention} shows that cross-attention from the ADM action expert to
WM subgoal-image tokens peaks during contact-establishment and placement
phases, confirming that forward scene predictions actively condition action
generation at critical manipulation moments.

\input{figures/fig_attention}

\subsection{Ablation Studies}

\cref{tab:ablation} ablates each major component introduced in
\cref{sec:training_stages}, building cumulatively on a $\pi_{0.5}$ model
trained from scratch on the same robot data.

\begin{table}[H]
\centering
\caption{\textbf{Component ablation.} Each row adds one module on top of the
  previous variant. Metrics: LIBERO-Long success (\%) and real-robot average
  success (\%).}
\label{tab:ablation}
\setlength{\tabcolsep}{3.5pt}
\renewcommand{\arraystretch}{1.15}
\small
\begin{tabular}{lcc}
\toprule
\textbf{Variant}
  & \textbf{LIBERO-Long} & \textbf{Real-robot} \\
\midrule
Base VLA ($\pi_{0.5}$ from scratch)   & 56.8 & 17.4 \\
+ DAT \& NTP (Stage~2)                  & 64.5 & 24.6 \\
+ ADM w/ isotropic noise              & 74.8 & 41.2 \\
+ Adaptive MoE                        & 84.0 & 55.8 \\
+ Historical-frame encoder            & 89.6 & 63.5 \\
+ WM subgoal injection (async)        & 93.8 & 70.8 \\
+ Covariance-correlated noise           & 96.4 & 75.2 \\
\rowcolor{primeblue!10}
\textbf{\method full (Stage~4 SFT)} & \textbf{98.6} & \textbf{78.1} \\
\bottomrule
\end{tabular}
\end{table}

Key findings:
\textbf{(i)~ADM and MoE.} Replacing the autoregressive head with ADM and adding
Adaptive MoE yields the largest early gain (+19.5\% on LIBERO-Long over
Stage~2), validating manifold-constrained flow matching with phase-specialised
experts.
\textbf{(ii)~Temporal conditioning.} The historical-frame encoder and WM
subgoal injection contribute complementary improvements (+5.6\% and +4.2\% on
LIBERO-Long, respectively): $\mathbf{W}_{hist}$ stabilises past context,
while $\widehat{\bo}_t^{\,(wm)}$ supplies forward dynamics anchors.
\textbf{(iii)~Covariance noise and SFT.} Empirical covariance noise
(+2.6\% LIBERO-Long) and Stage~4 joint fine-tuning (+2.2\%) provide consistent
final gains, indicating that kinematic-structure-aware sampling and unified
optimisation of $\cW$, VLM, and the action expert are both important for
deployment-ready performance.

\subsection{Scaling Analysis}

As shown in \cref{fig:scaling}, \method performance on LIBERO-Long scales
monotonically with world-model capacity, with the largest gain between Small
(5M) and Base (50M) parameters (+5.9\%). This suggests that subgoal-image
prediction quality is a key bottleneck: better video foundation models directly
improve the forward temporal anchors on which the VLM and ADM decoder rely,
without increasing inference latency thanks to asynchronous WM execution.

\input{figures/fig_scaling}

%% file: figures/fig_attention.tex
\begin{figure}[H]
  \centering
  \begin{tikzpicture}
  \begin{axis}[
    width=\linewidth,
    height=4.5cm,
    xlabel={Timestep},
    ylabel={Attn.\ to WM subgoal tokens},
    xmin=0,   xmax=30,
    ymin=0,   ymax=1,
    xtick={0,5,10,15,20,25,30},
    ytick={0,0.2,0.4,0.6,0.8,1.0},
    legend pos=north west,
    legend style={font=\scriptsize},
    grid=both, grid style={dotted, gray!40},
    tick label style={font=\scriptsize},
    label style={font=\small},
    mark size=0pt,
  ]
    \addplot[thick, color=primeblue, fill=primeblue!20]
      coordinates {
        (0,0.15)(2,0.18)(4,0.22)(6,0.30)(8,0.45)(10,0.72)(12,0.88)
        (14,0.65)(16,0.40)(18,0.35)(20,0.28)(22,0.62)(24,0.81)(26,0.75)
        (28,0.38)(30,0.20)
      } \closedcycle;
    \addlegendentry{WM subgoal-token attention (mean)}

    \draw[dashed, darkgray] (axis cs:7,0) -- (axis cs:7,1)
      node[above, font=\tiny, rotate=90, xshift=-10pt] {approach};
    \draw[dashed, darkgray] (axis cs:11,0) -- (axis cs:11,1)
      node[above, font=\tiny, rotate=90, xshift=-10pt] {contact};
    \draw[dashed, darkgray] (axis cs:21,0) -- (axis cs:21,1)
      node[above, font=\tiny, rotate=90, xshift=-10pt] {lift};
    \draw[dashed, darkgray] (axis cs:25,0) -- (axis cs:25,1)
      node[above, font=\tiny, rotate=90, xshift=-10pt] {place};
  \end{axis}
  \end{tikzpicture}
  \caption{\textbf{ADM cross-attention to WM subgoal-image tokens across
    manipulation phases.} Attention to WM-predicted subgoal tokens peaks at
    contact-establishment and placement phases, confirming that forward scene
    predictions actively condition manifold-constrained action generation during
    critical moments.}
  \label{fig:attention}
\end{figure}
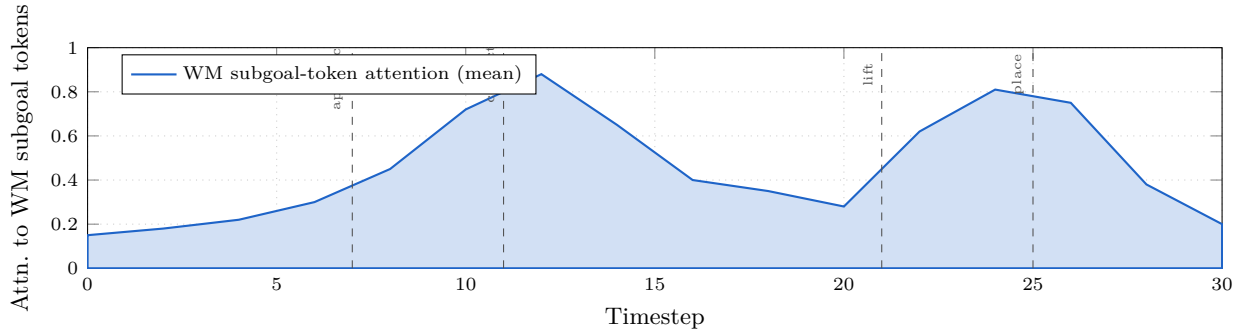

%% file: figures/fig_scaling.tex
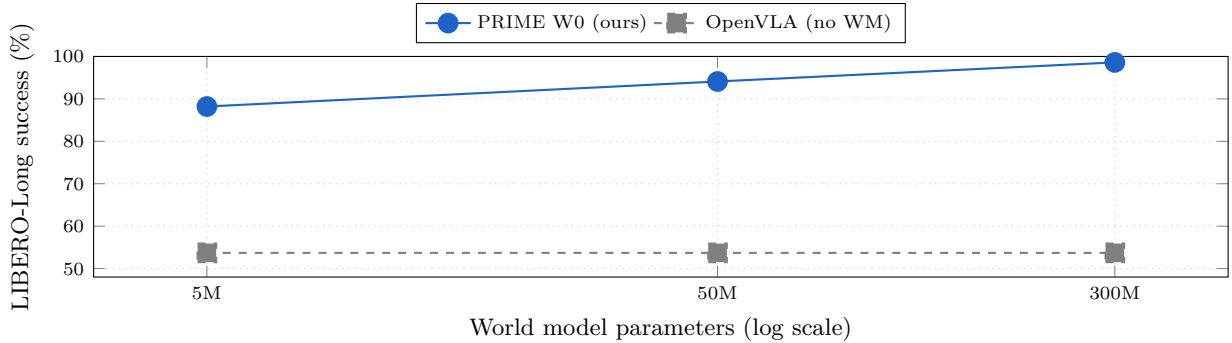
\begin{figure}[H]
  \centering
  \begin{tikzpicture}
  \begin{axis}[
    width=\linewidth,
    height=4.5cm,
    xlabel={World model parameters (log scale)},
    ylabel={LIBERO-Long success (\%)},
    xmode=log,
    xmin=3e6, xmax=5e8,
    ymin=48,  ymax=100,
    ytick={50,60,70,80,90,100},
    xtick={5e6,5e7,3e8},
    xticklabels={5M, 50M, 300M},
    legend style={
      font=\scriptsize,
      at={(0.5,1.06)},
      anchor=south,
      legend columns=2,
    },
    grid=both, grid style={dotted, gray!40},
    tick label style={font=\scriptsize},
    label style={font=\small},
    mark size=3.5pt,
  ]
    \addplot[thick, color=primeblue, mark=*]
      coordinates {(5e6,88.2)(5e7,94.1)(3e8,98.6)};
    \addlegendentry{\method (ours)}

    \addplot[thick, dashed, color=gray, mark=square*]
      coordinates {(5e6,53.7)(5e7,53.7)(3e8,53.7)};
    \addlegendentry{OpenVLA (no WM)}
  \end{axis}
  \end{tikzpicture}
  \caption{\textbf{Scaling with world-model capacity.} LIBERO-Long success
    scales monotonically with WM size; the largest gain is between 5M and
    50M parameters (+5.9\%). Better subgoal-image prediction directly benefits
    the asynchronous WM$\to$VLM$\to$ADM pipeline.}
  \label{fig:scaling}
\end{figure}

%% file: sections/05_discussion.tex
\section{Discussion}
\label{sec:discussion}

\paragraph{Why do temporal behaviours emerge?}
When the policy is conditioned on both historical tokens $\mathbf{W}_{hist}$
and WM-predicted subgoal images $\widehat{\bo}_t^{\,(wm)}$, the learned
mapping $\pi(\mathbf{A}_t \mid \bo_t, \ell, \mathbf{W}_{hist},
\widehat{\bo}_t^{\,(wm)})$ must remain consistent across varying temporal
contexts and forward scene predictions. This \emph{temporal consistency
constraint} discourages memorising isolated observation--action pairs and
encourages policies that track phase, anticipate contact, and modulate motion
dynamics. Behaviours such as subgoal-aligned control, reduced action jerk, and
phase-dependent MoE routing arise as efficient solutions to this constrained
learning problem rather than from task-specific supervision.

\paragraph{Relationship to dynamics-aware manipulation.}
\method{} represents a step toward \emph{dynamics-grounded generalist
manipulation}: policies that reason about \emph{how scenes evolve} rather than
mapping static appearances to actions. This is complementary to purely
feed-forward VLAs and to methods that use world models only as offline data
generators. By injecting explicit subgoal images asynchronously, \method{}
retains the interpretability of visual foresight while preserving real-time
deployability---the WM branch overlaps with current-frame processing rather
than serialising inference.

\paragraph{Limitations.}
\begin{enumerate}[leftmargin=1.5em]
  \item \emph{Subgoal-image fidelity}: ADM and the VLM depend on $\cW$
    producing visually coherent subgoal frames. For rapidly changing dynamics
    or scenes far from the WM training distribution, misaligned subgoals may
    mislead the policy.
  \item \emph{Historical window length}: the DINOv2 historical encoder captures
    a fixed temporal window $\tau$; very long-horizon tasks may still require
    explicit memory mechanisms beyond cached frames.
  \item \emph{MoE routing stability}: although temporal gating mitigates
    expert collapse, extremely imbalanced task distributions can still skew
    expert utilisation despite $\cL_{\mathrm{bal}}$.
  \item \emph{Tactile and force feedback}: \method{} infers dynamics from
    vision and proprioception only; integrating wrist force-torque or tactile
    sensing could improve contact-rich manipulation.
\end{enumerate}

\paragraph{Future directions.}
Promising extensions include end-to-end differentiable subgoal prediction,
scaling WM capacity with stronger video foundation models, combining
$\mathbf{W}_{hist}$ with learned episodic memory, and extending ADM to
whole-body humanoid platforms where kinematic coupling is even more pronounced.

%% file: sections/06_conclusion.tex
\section{Conclusion}
\label{sec:conclusion}

We introduced \method{}, a hierarchical conditional flow-matching VLA that
couples per-layer VLM KV injection, asynchronous world-model subgoal-image
prediction, historical-frame semantic bridging, and Adaptive MoE / ADM action
generation on a covariance-structured action manifold. A four-stage recipe
(\cref{fig:pretrain_pipeline}) progressively builds VLM grounding, DAT+NTP
visuomotor competence on a ${\sim}$10{,}000-hour teleop / UMI / simulation
mixture (\cref{fig:data_mixture}), a DiT-style flow-matching action expert
with frozen-backbone pre-training, and end-to-end SFT. At inference, WM
subgoal generation runs asynchronously from $\bo_{t-1}$, so the policy remains
real-time while still conditioning on explicit forward scene priors.

\method{} achieves state-of-the-art results on LIBERO and an in-house
real-robot suite. Ablations confirm that
DAT+NTP, manifold-constrained flow matching with correlation priors, WM
subgoal injection, historical bridging, and Adaptive MoE each contribute
measurably to long-horizon and contact-rich performance. We argue that closing
the loop between semantic reasoning, dynamics foresight, and
manifold-constrained control is a promising path toward robotic policies that
are simultaneously instruction-grounded, temporally coherent, and
physically legitimate.

\paragraph{Acknowledgements.}
We thank the Open X-Embodiment team for the robot demonstration dataset and
the open-source video foundation model community for pretrained world-model
and vision encoders used in our experiments.

%% file: sections/07_appendix.tex
\section{World Model Architecture}
\label{app:wm}

Our world model $\Phi_{wm}$ is a conditional image/video generator trained on
human-centric and robot manipulation footage. At inference it takes the
previous multi-view observation $\bo_{t-1}$, optional proprioceptive history,
and language instruction $\ell$, and predicts a subgoal RGB image
$\widehat{\bo}_t^{\,(wm)} \in \R^{H\times W\times 3}$ at the current
timestep (\cref{eq:wm_predict}). Architecturally, $\Phi_{wm}$ comprises a
spatio-temporal VAE encoder--decoder paired with a compact flow-based denoiser
(5M--300M parameters in our scaling study). The decoder outputs full visual
frames rather than abstract latent codes; frames are re-encoded by the shared
vision encoder $E_\phi$ before injection into the VLM token sequence
(\cref{eq:token_concat}). WM inference is launched asynchronously from
$\bo_{t-1}$ while $\bo_t$ is captured and prefilled, so end-to-end latency is
approximately $\max(T_{wm}, T_{vision})$.

\section{Historical Frame Encoder}
\label{app:hist}

The historical-frame encoder (\cref{sec:hist_encoder}; hyperparameters in
\cref{app:hist}) uses a frozen DINOv2
backbone $E_{\mathrm{dino}}$ to extract patch features from
$\{\bo_{t-\tau},\ldots,\bo_{t-1}\}$, followed by a two-layer projection with
LayerNorm to produce historical language tokens $\mathbf{W}_{hist}$
(\cref{eq:hist_bridge}). We use $\tau{=}8$ frames at the control frequency of
the target platform. The projection dimensions match the VLM embedding width
($d{=}4096$ in our implementation). Only the projection layers are trained;
DINOv2 weights remain frozen.

\section{Adaptive MoE and ADM Decoder}
\label{app:adm}

\paragraph{Adaptive MoE.}
Each of the 6 ADM decoder layers hosts $N{=}8$ lightweight FFN experts with
Top-$K{=}2$ activation. Selection and weighting are predicted by separate
sub-networks (\cref{eq:topk_select,eq:expert_weights}) with gating input
constructed from the current token representation and aggregated history
$\bar{\bh}_{t-\tau:t-1}$ (\cref{eq:gating_input}). Load-balancing loss
$\cL_{\mathrm{bal}}$ uses $\alpha_{\mathrm{bal}}{=}0.01$.

\paragraph{ADM flow matching.}
The action expert is a 6-layer transformer (512-dim hidden state, 8 attention
heads) with per-layer cross-attention to VLM KV representations
(\cref{eq:cross_attn}). Demonstration covariance $\widehat{\Sigma}_{\ba}$ is
estimated offline from 1M action chunks and factorised via Cholesky
decomposition (\cref{eq:corr_noise}). Inference uses 8-step Euler ODE
integration from $\mathcal{N}(0,\widehat{\Sigma}_{\ba})$ with action horizon
$H{=}16$ and execution window $h_{\mathrm{exec}}{=}8$.

%% file: references.bib
@article{rt2,
  author    = {Anthony Brohan and others},
  title     = {{RT-2}: Vision-Language-Action Models Transfer Web Knowledge
               to Robotic Control},
  journal   = {arXiv preprint arXiv:2307.15818},
  year      = {2023},
}

@inproceedings{openvla,
  author    = {Moo Jin Kim and others},
  title     = {{OpenVLA}: An Open-Source Vision-Language-Action Model},
  booktitle = {Proceedings of Machine Learning Research (PMLR)},
  year      = {2024},
  note      = {arXiv:2406.09246},
}

@article{pi0,
  author    = {Kevin Black and others},
  title     = {$\pi_0$: A Vision-Language-Action Flow Model},
  journal   = {arXiv preprint arXiv:2410.24164},
  year      = {2024},
}

@article{octo,
  author    = {Dibya Ghosh and others},
  title     = {Octo: An Open-Source Generalist Robot Policy},
  journal   = {arXiv preprint arXiv:2405.12213},
  year      = {2024},
}

@article{seer,
  author    = {Simeon Adebola and others},
  title     = {Seer: Predictive Inverse Dynamics Model for Robot Manipulation},
  journal   = {arXiv preprint arXiv:2412.15109},
  year      = {2024},
}

@article{vpp,
  title     = {Video Prediction Policy: A Video Prediction Model for
               Robotic Manipulation},
  journal   = {arXiv preprint arXiv:2412.14803},
  year      = {2024},
}

@misc{intelligence2026pi07steerablegeneralistrobotic,
      title={${\pi}_{0.7}$: a Steerable Generalist Robotic Foundation Model with Emergent Capabilities}, 
      author={Physical Intelligence and Bo Ai and Ali Amin and Raichelle Aniceto and Ashwin Balakrishna and Greg Balke and Kevin Black and George Bokinsky and Shihao Cao and Thomas Charbonnier and Vedant Choudhary and Foster Collins and Ken Conley and Grace Connors and James Darpinian and Karan Dhabalia and Maitrayee Dhaka and Jared DiCarlo and Danny Driess and Michael Equi and Adnan Esmail and Yunhao Fang and Chelsea Finn and Catherine Glossop and Thomas Godden and Ivan Goryachev and Lachlan Groom and Haroun Habeeb and Hunter Hancock and Karol Hausman and Gashon Hussein and Victor Hwang and Brian Ichter and Connor Jacobsen and Szymon Jakubczak and Rowan Jen and Tim Jones and Gregg Kammerer and Ben Katz and Liyiming Ke and Mairbek Khadikov and Chandra Kuchi and Marinda Lamb and Devin LeBlanc and Brendon LeCount and Sergey Levine and Xinyu Li and Adrian Li-Bell and Vladislav Lialin and Zhonglin Liang and Wallace Lim and Yao Lu and Enyu Luo and Vishnu Mano and Nandan Marwaha and Aikys Mongush and Liam Murphy and Suraj Nair and Tyler Patterson and Karl Pertsch and Allen Z. Ren and Gavin Schelske and Charvi Sharma and Baifeng Shi and Lucy Xiaoyang Shi and Laura Smith and Jost Tobias Springenberg and Kyle Stachowicz and Will Stoeckle and Jiaming Tang and Jimmy Tanner and Shalom Tekeste and Marcel Torne and Kyle Vedder and Quan Vuong and Anna Walling and Haohuan Wang and Jason Wang and XuDong Wang and Chris Whalen and Samuel Whitmore and Blake Williams and Charles Xu and Sukwon Yoo and Lili Yu and Wuming Zhang and Zhuoyang Zhang and Ury Zhilinsky},
      year={2026},
      eprint={2604.15483},
      archivePrefix={arXiv},
      primaryClass={cs.LG},
      url={https://arxiv.org/abs/2604.15483}, 
}

@misc{wu2026pragmaticvlafoundationmodel,
      title={A Pragmatic VLA Foundation Model}, 
      author={Wei Wu and Fan Lu and Yunnan Wang and Shuai Yang and Shi Liu and Fangjing Wang and Qian Zhu and He Sun and Yong Wang and Shuailei Ma and Yiyu Ren and Kejia Zhang and Hui Yu and Jingmei Zhao and Shuai Zhou and Zhenqi Qiu and Houlong Xiong and Ziyu Wang and Zechen Wang and Ran Cheng and Yong-Lu Li and Yongtao Huang and Xing Zhu and Yujun Shen and Kecheng Zheng},
      year={2026},
      eprint={2601.18692},
      archivePrefix={arXiv},
      primaryClass={cs.RO},
      url={https://arxiv.org/abs/2601.18692}, 
}

@misc{zhang2026joyaira01foundationmodel,
      title={JoyAI-RA 0.1: A Foundation Model for Robotic Autonomy}, 
      author={Tianle Zhang and Zhihao Yuan and Dafeng Chi and Peidong Liu and Dongwei Li and Kejun Hu and Likui Zhang and Junnan Nie and Ziming Wei and Zengjue Chen and Yili Tang and Jiayi Li and Zhiyuan Xiang and Mingyang Li and Tianci Luo and Hanwen Wan and Ao Li and Linbo Zhai and Zhihao Zhan and Xiaodong Bai and Jiakun Cai and Peng Cao and Kangliang Chen and Siang Chen and Yixiang Dai and Shuai Di and Yicheng Gong and Chenguang Gui and Yucheng Guo and Peng Hao and Qingrong He and Haoyang Huang and Kunrui Huang and Zhixuan Huang and Shibo Jin and Yixiang Jin and Anson Li and Dongjiang Li and Jiawei Li and Ruodai Li and Yihang Li and Yuzhen Li and Jiaming Liang and Fangsheng Liu and Jing Long and Mingxi Luo and Xing Pan and Hui Shen and Xiaomeng Tian and Daming Wang and Song Wang and Junwu Xiong and Hang Xu and Wanting Xu and Zhengcheng Yu and He Zhang and Jiyao Zhang and Lin Zhao and Chen Zhou and Nan Duan and Yuzheng Zhuang and Liang Lin},
      year={2026},
      eprint={2604.20100},
      archivePrefix={arXiv},
      primaryClass={cs.RO},
      url={https://arxiv.org/abs/2604.20100}, 
}

@misc{cai2026xiaomirobotics0opensourcedvisionlanguageactionmodel,
      title={Xiaomi-Robotics-0: An Open-Sourced Vision-Language-Action Model with Real-Time Execution}, 
      author={Rui Cai and Jun Guo and Xinze He and Piaopiao Jin and Jie Li and Bingxuan Lin and Futeng Liu and Wei Liu and Fei Ma and Kun Ma and Feng Qiu and Heng Qu and Yifei Su and Qiao Sun and Dong Wang and Donghao Wang and Yunhong Wang and Rujie Wu and Diyun Xiang and Yu Yang and Hangjun Ye and Yuan Zhang and Quanyun Zhou},
      year={2026},
      eprint={2602.12684},
      archivePrefix={arXiv},
      primaryClass={cs.RO},
      url={https://arxiv.org/abs/2602.12684}, 
}

@misc{robotics2026habilisbetafastmotionlonglastingondevice,
      title={Habilis-$\beta$: A Fast-Motion and Long-Lasting On-Device Vision-Language-Action Model}, 
      author={Tommoro Robotics and : and Jesoon Kang and Taegeon Park and Jisu An and Soo Min Kimm and Jaejoon Kim and Jinu Pahk and Byungju Kim and Junseok Lee and Namheon Baek and Sungwan Ha and Hojun Baek and Eduardo Ayerve Cruz and Wontae Kim and Junghyeon Choi and Yousuk Lee and Joonmo Han and Sunghyun Cho and Sunghyun Kwon and Soyoung Lee and Jun Ki Lee and Seung-Joon Yi and Byoung-Tak Zhang and Theo Taeyeong Kim},
      year={2026},
      eprint={2602.18813},
      archivePrefix={arXiv},
      primaryClass={cs.RO},
      url={https://arxiv.org/abs/2602.18813}, 
}

@misc{shou2026halounifiedvisionlanguageactionmodel,
      title={HALO: A Unified Vision-Language-Action Model for Embodied Multimodal Chain-of-Thought Reasoning}, 
      author={Quanxin Shou and Fangqi Zhu and Shawn Chen and Puxin Yan and Zhengyang Yan and Yikun Miao and Xiaoyi Pang and Zicong Hong and Ruikai Shi and Hao Huang and Jie Zhang and Song Guo},
      year={2026},
      eprint={2602.21157},
      archivePrefix={arXiv},
      primaryClass={cs.RO},
      url={https://arxiv.org/abs/2602.21157}, 
}

@misc{zheng2026pokevlaempoweringpocketsizedvisionlanguageaction,
      title={PokeVLA: Empowering Pocket-Sized Vision-Language-Action Model with Comprehensive World Knowledge Guidance}, 
      author={Yupeng Zheng and Xiang Li and Songen Gu and Yuhang Zheng and Shuai Tian and Weize Li and Linbo Wang and Senyu Fei and Pengfei Li and Yinfeng Gao and Zebin Xing and Yilun Chen and Qichao Zhang and Haoran Li and Wenchao Ding},
      year={2026},
      eprint={2604.20834},
      archivePrefix={arXiv},
      primaryClass={cs.RO},
      url={https://arxiv.org/abs/2604.20834}, 
}

@misc{gao2026dreamdojogeneralistrobotworld,
      title={DreamDojo: A Generalist Robot World Model from Large-Scale Human Videos}, 
      author={Shenyuan Gao and William Liang and Kaiyuan Zheng and Ayaan Malik and Seonghyeon Ye and Sihyun Yu and Wei-Cheng Tseng and Yuzhu Dong and Kaichun Mo and Chen-Hsuan Lin and Qianli Ma and Seungjun Nah and Loic Magne and Jiannan Xiang and Yuqi Xie and Ruijie Zheng and Dantong Niu and You Liang Tan and K. R. Zentner and George Kurian and Suneel Indupuru and Pooya Jannaty and Jinwei Gu and Jun Zhang and Jitendra Malik and Pieter Abbeel and Ming-Yu Liu and Yuke Zhu and Joel Jang and Linxi "Jim" Fan},
      year={2026},
      eprint={2602.06949},
      archivePrefix={arXiv},
      primaryClass={cs.RO},
      url={https://arxiv.org/abs/2602.06949}, 
}

@misc{ye2026worldactionmodelszeroshot,
      title={World Action Models are Zero-shot Policies}, 
      author={Seonghyeon Ye and Yunhao Ge and Kaiyuan Zheng and Shenyuan Gao and Sihyun Yu and George Kurian and Suneel Indupuru and You Liang Tan and Chuning Zhu and Jiannan Xiang and Ayaan Malik and Kyungmin Lee and William Liang and Nadun Ranawaka and Jiasheng Gu and Yinzhen Xu and Guanzhi Wang and Fengyuan Hu and Avnish Narayan and Johan Bjorck and Jing Wang and Gwanghyun Kim and Dantong Niu and Ruijie Zheng and Yuqi Xie and Jimmy Wu and Qi Wang and Ryan Julian and Danfei Xu and Yilun Du and Yevgen Chebotar and Scott Reed and Jan Kautz and Yuke Zhu and Linxi "Jim" Fan and Joel Jang},
      year={2026},
      eprint={2602.15922},
      archivePrefix={arXiv},
      primaryClass={cs.RO},
      url={https://arxiv.org/abs/2602.15922}, 
}

@misc{yuan2026fastwamworldactionmodels,
      title={Fast-WAM: Do World Action Models Need Test-time Future Imagination?}, 
      author={Tianyuan Yuan and Zibin Dong and Yicheng Liu and Hang Zhao},
      year={2026},
      eprint={2603.16666},
      archivePrefix={arXiv},
      primaryClass={cs.CV},
      url={https://arxiv.org/abs/2603.16666}, 
}

@misc{ma2026dit4ditjointlymodelingvideo,
      title={DiT4DiT: Jointly Modeling Video Dynamics and Actions for Generalizable Robot Control}, 
      author={Teli Ma and Jia Zheng and Zifan Wang and Chunli Jiang and Andy Cui and Junwei Liang and Shuo Yang},
      year={2026},
      eprint={2603.10448},
      archivePrefix={arXiv},
      primaryClass={cs.RO},
      url={https://arxiv.org/abs/2603.10448}, 
}

@misc{luo2026beingh07latentworldactionmodel,
      title={Being-H0.7: A Latent World-Action Model from Egocentric Videos}, 
      author={Hao Luo and Wanpeng Zhang and Yicheng Feng and Sipeng Zheng and Haiweng Xu and Chaoyi Xu and Ziheng Xi and Yuhui Fu and Zongqing Lu},
      year={2026},
      eprint={2605.00078},
      archivePrefix={arXiv},
      primaryClass={cs.RO},
      url={https://arxiv.org/abs/2605.00078}, 
}

@misc{bi2025motusunifiedlatentaction,
      title={Motus: A Unified Latent Action World Model}, 
      author={Hongzhe Bi and Hengkai Tan and Shenghao Xie and Zeyuan Wang and Shuhe Huang and Haitian Liu and Ruowen Zhao and Yao Feng and Chendong Xiang and Yinze Rong and Hongyan Zhao and Hanyu Liu and Zhizhong Su and Lei Ma and Hang Su and Jun Zhu},
      year={2025},
      eprint={2512.13030},
      archivePrefix={arXiv},
      primaryClass={cs.CV},
      url={https://arxiv.org/abs/2512.13030}, 
}

@misc{liao2025genieenvisionerunifiedworld,
      title={Genie Envisioner: A Unified World Foundation Platform for Robotic Manipulation}, 
      author={Yue Liao and Pengfei Zhou and Siyuan Huang and Donglin Yang and Shengcong Chen and Yuxin Jiang and Yue Hu and Jingbin Cai and Si Liu and Jianlan Luo and Liliang Chen and Shuicheng Yan and Maoqing Yao and Guanghui Ren},
      year={2025},
      eprint={2508.05635},
      archivePrefix={arXiv},
      primaryClass={cs.RO},
      url={https://arxiv.org/abs/2508.05635}, 
}

@misc{ye2026gigaworldpolicyefficientactioncenteredworldaction,
      title={GigaWorld-Policy: An Efficient Action-Centered World--Action Model}, 
      author={Angen Ye and Boyuan Wang and Chaojun Ni and Guan Huang and Guosheng Zhao and Hao Li and Hengtao Li and Jie Li and Jindi Lv and Jingyu Liu and Min Cao and Peng Li and Qiuping Deng and Wenjun Mei and Xiaofeng Wang and Xinze Chen and Xinyu Zhou and Yang Wang and Yifan Chang and Yifan Li and Yukun Zhou and Yun Ye and Zhichao Liu and Zheng Zhu},
      year={2026},
      eprint={2603.17240},
      archivePrefix={arXiv},
      primaryClass={cs.CV},
      url={https://arxiv.org/abs/2603.17240}, 
}

@misc{kim2026cosmospolicyfinetuningvideo,
      title={Cosmos Policy: Fine-Tuning Video Models for Visuomotor Control and Planning}, 
      author={Moo Jin Kim and Yihuai Gao and Tsung-Yi Lin and Yen-Chen Lin and Yunhao Ge and Grace Lam and Percy Liang and Shuran Song and Ming-Yu Liu and Chelsea Finn and Jinwei Gu},
      year={2026},
      eprint={2601.16163},
      archivePrefix={arXiv},
      primaryClass={cs.AI},
      url={https://arxiv.org/abs/2601.16163}, 
}

@misc{robbyantteam2026advancingopensourceworldmodels,
      title={Advancing Open-source World Models}, 
      author={Robbyant Team and Zelin Gao and Qiuyu Wang and Yanhong Zeng and Jiapeng Zhu and Ka Leong Cheng and Yixuan Li and Hanlin Wang and Yinghao Xu and Shuailei Ma and Yihang Chen and Jie Liu and Yansong Cheng and Yao Yao and Jiayi Zhu and Yihao Meng and Kecheng Zheng and Qingyan Bai and Jingye Chen and Zehong Shen and Yue Yu and Xing Zhu and Yujun Shen and Hao Ouyang},
      year={2026},
      eprint={2601.20540},
      archivePrefix={arXiv},
      primaryClass={cs.CV},
      url={https://arxiv.org/abs/2601.20540}, 
}

@misc{motubrainteam2026motubrainadvancedworldaction,
      title={MotuBrain: An Advanced World Action Model for Robot Control}, 
      author={MotuBrain Team and Chendong Xiang and Fan Bao and Haitian Liu and Hengkai Tan and Hongzhe Bi and James Li and Jiabao Liu and Jingrui Pang and Kiro Jing and Louis Liu and Mengchen Cai and Rongxu Cui and Ruowen Zhao and Runqing Wang and Shuhe Huang and Yao Feng and Yinze Rong and Zeyuan Wang and Jun Zhu},
      year={2026},
      eprint={2604.27792},
      archivePrefix={arXiv},
      primaryClass={cs.RO},
      url={https://arxiv.org/abs/2604.27792}, 
}

@misc{assran2025vjepa2selfsupervisedvideo,
      title={V-JEPA 2: Self-Supervised Video Models Enable Understanding, Prediction and Planning}, 
      author={Mido Assran and Adrien Bardes and David Fan and Quentin Garrido and Russell Howes and Mojtaba and Komeili and Matthew Muckley and Ammar Rizvi and Claire Roberts and Koustuv Sinha and Artem Zholus and Sergio Arnaud and Abha Gejji and Ada Martin and Francois Robert Hogan and Daniel Dugas and Piotr Bojanowski and Vasil Khalidov and Patrick Labatut and Francisco Massa and Marc Szafraniec and Kapil Krishnakumar and Yong Li and Xiaodong Ma and Sarath Chandar and Franziska Meier and Yann LeCun and Michael Rabbat and Nicolas Ballas},
      year={2025},
      eprint={2506.09985},
      archivePrefix={arXiv},
      primaryClass={cs.AI},
      url={https://arxiv.org/abs/2506.09985}, 
}

@misc{cen2025worldvlaautoregressiveactionworld,
      title={WorldVLA: Towards Autoregressive Action World Model}, 
      author={Jun Cen and Chaohui Yu and Hangjie Yuan and Yuming Jiang and Siteng Huang and Jiayan Guo and Xin Li and Yibing Song and Hao Luo and Fan Wang and Deli Zhao and Hao Chen},
      year={2025},
      eprint={2506.21539},
      archivePrefix={arXiv},
      primaryClass={cs.RO},
      url={https://arxiv.org/abs/2506.21539}, 
}

@misc{cen2025rynnvla002unifiedvisionlanguageactionworld,
      title={RynnVLA-002: A Unified Vision-Language-Action and World Model}, 
      author={Jun Cen and Siteng Huang and Yuqian Yuan and Kehan Li and Hangjie Yuan and Chaohui Yu and Yuming Jiang and Jiayan Guo and Xin Li and Hao Luo and Fan Wang and Deli Zhao and Hao Chen},
      year={2025},
      eprint={2511.17502},
      archivePrefix={arXiv},
      primaryClass={cs.RO},
      url={https://arxiv.org/abs/2511.17502}, 
}

@misc{gigabrainteam2025gigabrain0worldmodelpoweredvisionlanguageaction,
      title={GigaBrain-0: A World Model-Powered Vision-Language-Action Model}, 
      author={GigaBrain Team and Angen Ye and Boyuan Wang and Chaojun Ni and Guan Huang and Guosheng Zhao and Haoyun Li and Jie Li and Jiagang Zhu and Lv Feng and Peng Li and Qiuping Deng and Runqi Ouyang and Wenkang Qin and Xinze Chen and Xiaofeng Wang and Yang Wang and Yifan Li and Yilong Li and Yiran Ding and Yuan Xu and Yun Ye and Yukun Zhou and Zhehao Dong and Zhenan Wang and Zhichao Liu and Zheng Zhu},
      year={2025},
      eprint={2510.19430},
      archivePrefix={arXiv},
      primaryClass={cs.RO},
      url={https://arxiv.org/abs/2510.19430}, 
}

@misc{hung2025nora15visionlanguageactionmodeltrained,
      title={NORA-1.5: A Vision-Language-Action Model Trained using World Model- and Action-based Preference Rewards}, 
      author={Chia-Yu Hung and Navonil Majumder and Haoyuan Deng and Liu Renhang and Yankang Ang and Amir Zadeh and Chuan Li and Dorien Herremans and Ziwei Wang and Soujanya Poria},
      year={2025},
      eprint={2511.14659},
      archivePrefix={arXiv},
      primaryClass={cs.RO},
      url={https://arxiv.org/abs/2511.14659}, 
}

@misc{li2025vlarftvisionlanguageactionreinforcementfinetuning,
      title={VLA-RFT: Vision-Language-Action Reinforcement Fine-tuning with Verified Rewards in World Simulators}, 
      author={Hengtao Li and Pengxiang Ding and Runze Suo and Yihao Wang and Zirui Ge and Dongyuan Zang and Kexian Yu and Mingyang Sun and Hongyin Zhang and Donglin Wang and Weihua Su},
      year={2025},
      eprint={2510.00406},
      archivePrefix={arXiv},
      primaryClass={cs.RO},
      url={https://arxiv.org/abs/2510.00406}, 
}

@misc{liu2026worldvlaloopclosedlooplearningvideo,
      title={World-VLA-Loop: Closed-Loop Learning of Video World Model and VLA Policy}, 
      author={Xiaokang Liu and Zechen Bai and Hai Ci and Kevin Yuchen Ma and Mike Zheng Shou},
      year={2026},
      eprint={2602.06508},
      archivePrefix={arXiv},
      primaryClass={cs.RO},
      url={https://arxiv.org/abs/2602.06508}, 
}

@misc{zhang2026practicalworldmodelbasedreinforcement,
      title={Towards Practical World Model-based Reinforcement Learning for Vision-Language-Action Models}, 
      author={Zhilong Zhang and Haoxiang Ren and Yihao Sun and Yifei Sheng and Haonan Wang and Haoxin Lin and Zhichao Wu and Pierre-Luc Bacon and Yang Yu},
      year={2026},
      eprint={2603.20607},
      archivePrefix={arXiv},
      primaryClass={cs.RO},
      url={https://arxiv.org/abs/2603.20607}, 
}

@article{pertsch2025fast,
  title     = {{FAST}: Efficient Action Tokenization for
               Vision-Language-Action Models},
  author    = {Pertsch, Karl and Stachowicz, Kyle and Ichter, Brian
               and Driess, Danny and Nair, Suraj and Vuong, Quan
               and Mees, Oier and Finn, Chelsea and Levine, Sergey},
  journal   = {arXiv preprint arXiv:2501.09747},
  year      = {2025}
}

@inproceedings{bjorck2025grootn1,
  title     = {{GR00T N1}: An Open Foundation Model for
               Generalist Humanoid Robots},
  author    = {NVIDIA and Bjorck, Johan and Casta{\~n}eda, Fernando
               and Cherniadev, Nikita and Da, Xingye and Ding, Runyu
               and Fan, Linxi and Fang, Yu and Fox, Dieter
               and Hu, Fengyuan and Huang, Spencer and Jang, Joel
               and Jiang, Zhenyu and Kautz, Jan and others},
  booktitle = {arXiv Preprint},
  eprint    = {2503.14734},
  year      = {2025}
}

@article{bu2025univla,
  title     = {{UniVLA}: Learning to Act Anywhere with
               Task-centric Latent Actions},
  author    = {Bu, Qingwen and Yang, Yanting and Cai, Jisong
               and Gao, Shenyuan and Ren, Guanghui and Yao, Maoqing
               and Luo, Ping and Li, Hongyang},
  journal   = {arXiv preprint arXiv:2505.06111},
  year      = {2025}
}

@article{liang2025ddvla,
  title     = {Discrete Diffusion {VLA}: Bringing Discrete Diffusion
               to Action Decoding in Vision-Language-Action Policies},
  author    = {Liang, Zhixuan and Li, Yizhuo and Yang, Tianshuo
               and Wu, Chengyue and Mao, Sitong and Pei, Liuao
               and Yang, Xiaokang and Pang, Jiangmiao
               and Mu, Yao and Luo, Ping},
  journal   = {arXiv preprint arXiv:2508.20072},
  year      = {2025}
}

@article{kim2025openvlaoft,
  title     = {Fine-Tuning Vision-Language-Action Models:
               Optimizing Speed and Success},
  author    = {Kim, Moo Jin and Finn, Chelsea and Liang, Percy},
  journal   = {arXiv preprint arXiv:2502.19645},
  year      = {2025}
}

@article{shi2025memoryvla,
  title     = {{MemoryVLA}: Perceptual-Cognitive Memory in
               Vision-Language-Action Models for Robotic Manipulation},
  author    = {Shi, Hao and Xie, Bin and Liu, Yingfei and Sun, Lin
               and Liu, Fengrong and Wang, Tiancai and Zhou, Erjin
               and Fan, Haoqiang and Zhang, Xiangyu and Huang, Gao},
  journal   = {arXiv preprint arXiv:2508.19236},
  year      = {2025}
}

@inproceedings{black2025pi05,
  title     = {$\pi_{0.5}$: A Vision-Language-Action Model with
               Open-World Generalization},
  author    = {{Physical Intelligence} and Black, Kevin and Brown, Noah
               and Darpinian, James and Dhabalia, Karan and Driess, Danny
               and Esmail, Adnan and Equi, Michael and Finn, Chelsea
               and Fusai, Niccolo and Galliker, Manuel Y.
               and Ghosh, Dibya and Groom, Lachy and Hausman, Karol
               and Ichter, Brian and others},
  booktitle = {Proceedings of The 9th Conference on Robot Learning
               (CoRL)},
  series    = {Proceedings of Machine Learning Research},
  volume    = {305},
  pages     = {17--40},
  year      = {2025}
}

@inproceedings{reuss2025flower,
  title     = {{FLOWER}: Democratizing Generalist Robot Policies with
               Efficient Vision-Language-Flow Models},
  author    = {Reuss, Moritz and Zhou, Hongyi and R{\"u}hle, Marcel
               and Ya{\u{g}}murlu, {\"O}mer Erdin{\c{c}}
               and Otto, Fabian and Lioutikov, Rudolf},
  booktitle = {9th Annual Conference on Robot Learning (CoRL)},
  year      = {2025},
  url       = {https://openreview.net/forum?id=JeppaebLRD}
}

@article{qu2025eo1,
  title     = {{EmbodiedOneVision}: Interleaved Vision-Text-Action
               Pretraining for General Robot Control},
  author    = {Qu, Delin and Song, Haoming and Chen, Qizhi
               and Chen, Zhaoqing and Gao, Xianqiang and Ye, Xinyi
               and Lv, Qi and Shi, Modi and Ren, Guanghui
               and Ruan, Cheng and Yao, Maoqing and Yang, Haoran
               and Bao, Jiacheng and Zhao, Bin and Wang, Dong},
  journal   = {arXiv preprint arXiv:2508.21112},
  year      = {2025}
}
